# 3D LULC classification using multispectral LiDAR and deep learning: current and prospective schemes

Narges Takhtkeshha[1,2]*, Aldino Rizaldy[3,4], Markus Hollaus[2], Juha Hyyppä[5], Fabio Remondino[1], Gottfried Mandlburger[2]

[1] 3D Optical Metrology (3DOM) Unit, Bruno Kessler Foundation (FBK), Trento, Italy; ntakhtkeshha@fbk.eu, remondino@fbk.eu
[2] Department of Geodesy and Geoinformation, TU Wien, Vienna, Austria; narges.takhtkeshha@geo.tuwien.ac.at, gottfried.mandlburger@geo.tuwien.ac.at, markus.hollaus@geo.tuwien.ac.at
[3] Helmholtz-ZentrumDresden-Rossendorf (HZDR), Helmholtz Institute Freiberg for Resource Technology (HIF), Freiberg, Germany; a.rizaldy@hzdr.de
[4] Freie Universität Berlin, Remote Sensing and Geoinformatics, Berlin, Germany; aldino.rizaldy@fu-berlin.de
[5] Department of Remote Sensing and Photogrammetry, Finnish Geospatial Research Institute FGI, The National Land Survey of Finland, Vuorimiehentie 5, Espoo, FI-02150, Finland; juha.hyyppa@nls.fi

## Abstract

Land Use Land Cover (LULC) classification is fundamental to national 3D mapping programs and geospatial analyses, resource management and sustainable development planning. While multispectral (MS) LiDAR (Light Detection and Ranging) technology offers synchronized spatial-spectral information and Deep Learning (DL) has shown promise in 3D point cloud semantic segmentation, significant barriers remain. Specifically, the lack of publicly available urban and suburban MS LiDAR datasets tailored to National Mapping and Cadastral Agencies (NMCAs)' LULC classification schemes, along with the absence of a consistent class definition scheme tailored to NMCAs schemes, limits the practical adoption of these technologies in operational workflows. Therefore, this study addresses these gaps by reporting NMCAs' current (L1) and prospective (L2) LULC classification schemes, presenting a novel LiDAR benchmark dataset at two levels of detail (L1 and L2), evaluating seven State-of-the-Art (SOTA) DL models, and finally conducting rigorous spectral ablation analyses at the two levels of detail. Benchmarking results demonstrate the superiority of the Point Transformer V3 model at both levels, achieving high mean Intersection over Union (mIoU) values of 79.4% for LULC-L1 (eight classes) and 58.9% (20 classes) for LULC-L2 by leveraging a two-wavelength commercial MS LiDAR system (532 nm and 1064 nm). Furthermore, spectral ablation studies highlight the significant potential of MS LiDAR data over purely geometric features. While the spectral advantage is marginal at LULC-L1 (a 1.1 Percentage Point (PP) improvement in mIoU), it becomes considerably more pronounced at LULC-L2 (a 7.8 pp improvement). Our findings underscore the capacity of LiDAR reflectance and MS LiDAR for surface material identification, supporting the prospective transition of NMCAs schemes toward fine-grained LULC classification. Additionally, the Loosdorf-MSL dataset serves as a valuable resource for improving consistency across national and international LULC products.

## 1. Introduction

Land Use and Land Cover (LULC) products are increasingly evolving towards finer-scale classification levels and higher spatial and temporal resolutions. This shift is driven by advances in Deep Learning (DL) and by the integration of multiple data sources (Prasad et al., 2020; Bo et al., 2024; Rizaldy et al., 2025). A transformative development in this domain is the advent of multispectral (MS) Light Detection and Ranging (LiDAR) systems. Unlike conventional LiDAR, which is primarily limited to geometric information and one spectral wavelength, MS LiDAR captures 3D data across multiple spectral wavelengths, enabling the simultaneous acquisition of spatial and spectral information (Kaasalainen et al., 2007; Riveiro et al., 2019; Takhtkeshha et al., 2024b). This spatial-spectral synergy provides a large number of spatial and spectral features for target classification, thereby improving the segmentation and classification of point clouds, as well as the quality of the obtained elevation models. Furthermore, MS LiDAR data facilitates fine-grained 3D mapping and single-source data generation, offering significant advantages for modelling complex urban scenes with diverse object types.

While current 3D urban mapping often relies on photogrammetric point clouds (Bayrak et al., 2023; Gao et al., 2024) or multimodal data fusion (Rizaldy et al., 2025; Wang et al., 2025b), these methods face inherent limitations. Photogrammetry is susceptible to occlusions and shadows, whereas data fusion of LiDAR with multi/hyperspectral imagery introduces significant operational complexities, including registration errors, temporal discrepancies (e.g., moving objects), and increased costs. In contrast, MS LiDAR provides shadow-free acquisition, superior height accuracy, and enhanced vegetation penetration within a single, synchronized framework.

Up to now, MS LiDAR data have been studied in a wide range of applications, including ecology and forestry, LULC classification, navigation, change detection, archaeology, geology, bathymetry, autonomous driving, etc. (Taher et al., 2022; Takhtkeshha et al., 2024b; Wu et al., 2024a; Zhou et al., 2025a; Takhtkeshha et al., 2025a; Ruoppa et al., 2025; Taher et al., 2026). Slightly after ecology and forestry, LULC classification is the second mostly used application of MS LiDAR data, covering 39.3% of the studies (Chen, 2018; Pan et al., 2020; Zhang et al., 2022; Takhtkeshha et al., 2024b; Reichler et al., 2024; Zhou et al., 2025a).

It has been proven that processing point clouds directly maximizes the advantages of 3D data and achieves more accurate results compared to projecting them onto images and applying image processing techniques (Tychola et al., 2024). Nevertheless, the irregular distribution of individual points of a point cloud, along with the substantial data volume of large-area 3D datasets, presents significant challenges for point-wise processing. This issue becomes even more pronounced when dealing with MS point clouds. Advanced DL models have attracted considerable attention for processing point clouds and have demonstrated remarkable potential (Zhou et al., 2025b). Ongoing progress in the development of new 3D DL algorithms encourage their application to MS point cloud processing. However, adapting 3D DL models to MS point clouds remains an active area of research (Takhtkeshha et al., 2025b; Ruoppa et al., 2026). Consequently, this paper focuses on processing suburban/urban MS LiDAR data for LULC classification using State-of-the-Art (SOTA) 3D DL models. Table 1 provides a detailed summary of studies employing MS LiDAR data and DL models for 3D LULC classification.

**Table 1** Overview of DL-based LULC classification studies using MS LiDAR data. For studies based on more than one model, the best-performing DL model is highlighted in bold.

| References | Data source(s) | Classes | Methodology(s) | MS effect w.r.t the geometry | Remarks |
|---|---|---|---|---|---|
| Zhou et al., 2025a | Optech Titan, ALS | 7: impervious ground, grass, tree, building, car, power line, and bare land | $^{3D}$ PointNet++, DGCNN, RandLA-Net, KPConv, **PAConv**, PTv1 | NA | New spectral reconstruction method (DossaNet) |
| Wang et al., 2025a | Optech Titan, ALS | 9: barren, building, car, grass, powerline, road, ship, tree, and water | $^{2D}$ MGC2N | NA | NA |
| Reichler et al., 2024 | FGI AkhkaR4-DW, backpack | 13: asphalt, soft ground, brick paving, building details, brick wall, vegetation, plastered wall, other man-made objects, curb, concrete wall, car, road marking, and noise | $^{2D}$ Encoder - decoder CNN | 18.1 pp in mIoU | More MS improvement over NIR channel<br>Wall material detection |
| Wang et al., 2024 | Optech Titan, ALS | 9: barren, car, powerline, ship, water, building, grass, road, and tree | $^{3D}$ MKGSL | 5.2 pp in mIoU | Empirical constraints to guide graph optimization |
| Takhtkeshha et al., 2024a | HeliALS-TW, ALS | 7: tree, roof, facade, car, grass, asphalt, and gravel | $^{3D}$ KPConv | NA | Automatically annotated data |
| Oinonen et al., 2024 | HeliALS-TW, ALS | 7: building, high vegetation, low vegetation, asphalt, rock, football field, and gravel | $^{3D}$ GroupSP | 24.9 pp in mIoU | Unsupervised |
| Chen et al., 2024 | Optech Titan, ALS | 6: road, building, land/grass, tree, water/soil, and powerline | $^{3D}$ FPWS-Net | NA | Weakly supervised |
| Yang et al., 2024 | Optech Titan, ALS | 6: road, grass, building, tree, car, and powerline | $^{3D}$ RF, PointNet, PointNet++, RandLA-Net, DGCNN, and **PTv1** | 2-5 pp in mIoU | NA |
| Xiao et al., 2022 | Optech Titan, ALS | 6: road, building, grass, tree, open ground, and powerline | 3D Improved RandLA-Net | NA | Improvement by redundancy reduction of the point cloud feature distribution using singular value decomposition |
| Zhang et al., 2022 | Optech Titan, ALS | 6: road, building, grass, tree, soil, and powerline | 3D MPT + network | 18.9 pp in mIoU | More MS improvement over SWIR channel |
| Li et al., 2022 | Optech Titan, ALS | 4: road, grass, tree, building | 3D AGFP-Net | 18.9 pp in KC | NA |
| Zhao et al., 2021b | Optech Titan, ALS | 6: tree, grass, road, soil, building, and powerline | 3D FR-GCNet | 16.1 pp in mIoU | More MS improvement over green channel |
| Jing et al., 2021 | Optech Titan, ALS | 6: road, building, grass, tree, soil, and powerline | 3D SE PointNet++ | 10.6 pp in mIoU | More MS improvement over green channel |
| Pan et al., 2020 | Optech Titan, ALS | 6: building, tree, road, grass, soil, and water | 2D 3D CNN | 4 pp in KC | NA |
| Pan et al., 2019 | Optech Titan, ALS | 6: building, tree, road, grass, soil, and water | 2D Deep Boltzmann machine | NA | NA |
| Chen, 2018 | Optech Titan, ALS | 6: water, tree, building, road, bare land, and other impervious surfaces | 2D 3D CNN | NA | NA |

According to Table 1, generally three MS LiDAR systems have been utilized in the literature for this purpose: (i) the Optech Titan (Chen, 2018; Jing et al., 2021; Li et al., 2022; Chen et al., 2024; Yang et al., 2024; Zhou et al., 2025a; Wang et al., 2025a), an Airborne Laser Scanner (ALS) launched in 2014 by Teledyne Optech (Vaughan, ON, Canada), operating at wavelengths of 1550 nm, 1064 nm, and 532 nm; (ii) the HeliALS-TW (1550 nm, 905 nm, and 532 nm; Takhtkeshha et al., 2024ba; Oinonen et al., 2024), developed by the Finnish Geospatial Research Institute (FGI), comprising three *RIEGL* single-wavelength scanners; (iii) the FGI AkhkaR4-DW (Reichler et al., 2024), a dual-wavelength (1550 nm and 905 nm) backpack laser scanner designed by FGI. According to literature findings, the spatial-spectral information captured by MS LiDAR systems enhances LULC classification, achieving up to a 24.9 Percentage Points (pp) improvement in mean Intersection over Union (mIoU; Oinonen et al., 2024). Noticeably, MS LiDAR systems facilitate distinguishing between objects with similar geometries but different radiometric signature. For instance, ground-level objects such as soil, road, road marking, asphalt, football field, rock, and gravel can be effectively differentiated using MS LiDAR data. Such systems also enable identification of various materials in off-ground objects (Reichler et al., 2024). As shown in Table 1, early studies predominantly rely on 2D approaches (Chen, 2018; Pan et al., 2019; Pan et al., 2020; Reichler et al., 2024; Wang et al., 2025a), whereas 3D DL models have become increasingly prevalent in recent research (Xiao et al., 2022; Zhang et al., 2022; Li et al., 2022; Takhtkeshha et al., 2024a; Chen et al., 2024; Oinonen et al., 2024, Wang et al., 2024, Zhou et al., 2025a). To date, several advanced DL architectures have been applied, including PointNet (Qi et al., 2017a; Yang et al., 2024), PointNet++ (Qi et al., 2017b; Yang et al., 2024; Zhou et al., 2025a), 3D Convolutional Neural Networks (CNN, Chen, 2018; Pan et al., 2020), deep Boltzmann machines (Pan et al., 2019), Squeeze-and-Excitation (SE) PointNet++ (Jing et al., 2021), FR-GCNet (Zhao et al., 2021b), Attentive Geometric Feature Pyramid Network (AGFP-Net; Li et al., 2022), MPT+ network (Zhang et al., 2022), RandLA-Net (Hu et al., 2020; Yang et al., 2024; Zhou et al., 2025a), improved RandLA-Net (Xiao et al., 2022), Dynamic Graph Convolutional Neural Network (DGCNN; Wang et al., 2019; Yang et al., 2024; Zhou et al., 2025a), Point Transformer V1 (PTv1; Zhao et al., 2021a; Yang et al., 2024; Zhou et al., 2025a), GroupSP (Oinonen et al., 2024), Multi-Kernel Graph Structure Learning (MKGSL; Wang et al., 2024), and Kernel Point Convolution (KPConv; Thomas et al., 2019; Takhtkeshha et al., 2024a; Zhou et al., 2025a). All existing studies, except for Takhtkeshha et al. (2024a) and Oinonen et al. (2024), rely on supervised methods.

By introducing AGFP-Net, Li et al. (2022) addressed the lack of comprehensive extraction of geometric structure features and the fusion of multi-scale features, improved the Kappa Coefficient (KP) by 18.9 pp. Yang et al. (2024) compared five DL models (PointNet, PointNet++, RandLA-Net, DGCNN, and PTv1) and a Random Forest (RF) machine learning model for LULC mapping across six classes: road, grass, building, tree, car, and powerline. Their study shows the superiority of PTv1 and furthermore highlights the advantage of the three-wavelength Optech Titan MS LiDAR system, which improve mIoU by 2–5 pp. Wang et al. (2024) proposed MKGSL method for classifying urban MS point clouds in nine classes. This method first captures adaptive multi-scale relationships within the MS point cloud by learning combinations of graphs. Their method then refines these graphs by incorporating a series of empirical a-priori constraints, such as low-rank and sparsity constraints, to guide the evolution of the graph structure. Reichler et al. (2024) conducted pioneering research utilizing AkhkaR4-DW to classify LULC in 13 classes (including wall materials). More recently, Wang et al. (2025a) proposed a 2D Masking Graph Cross-Convolution Network (MGC2N) for LULC classification using MS LiDAR data across nine classes: barren, building, car, grass, powerline, road, ship, tree, and water.

The proposed MGC2N incorporates a self-attention masking module and a spatial–spectral cross-convolution module, designed to establish point-to-point relationships independent of spatial distance. Recently, Zhou et al. (2025) introduced a new spectral reconstruction method that leverages a Dual-attention Spectral Optimization Reconstruction Network (DossaNet) to generate MS point clouds from individual LiDAR channels. They applied six DL models, namely PointNet++, DGCNN, RandLA-Net, KPConv, PAConv and PTv1, and demonstrated that DossaNet-based MS point clouds generation improves classification accuracy for most of these models compared to the inverse distance weighted and k-nearest neighbor approaches.

In the last decade, weakly supervised and unsupervised LULC classification approaches have gained attention in a few studies. Chen et al. (2024) proposed FPWS-Net, a novel weakly supervised DL framework for MS LiDAR point cloud classification. It employs a dual semantic inference structure and the feature perturbance generated by them were minimized by considering a consistency constraint loss and a mutual pseudo-labeling loss to enhance model stability. The integration of KPConv allows effective extraction of geometric and spectral features in a high-dimensional space, improving training supervision. Despite achieving strong performance with only 0.1% labeled data, the model struggled with certain classes, such as soil, due to limited supervised signals. Takhtkeshha et al. (2024a) proposed an unsupervised methodology based on automatically annotated data generated using the Segment Anything Model (SAM) and K-means clustering across seven classes, which were then used to train a KPConv DL model. Oinonen et al. (2024) classified urban areas into seven classes using a novel unsupervised DL method called GroupSP, inspired by the GrowSP (Zhang et al., 2023) algorithm. Their study substantiated that the three-wavelength HeliALS data boosted LULC classification by 24.9 pp in mIoU compared to only using geometric information.

With the advancement of data collection systems, either through multimodal remote sensing or single-source MS LiDAR systems, new DL models have emerged to process MS point clouds. Recently, Rizaldy et al. (2025) proposed a fully 3D DL model tailored for processing MS point clouds derived by fusion of MS images and conventional monochromatic LiDAR data, called HyperPointFormer (HPF). The HPF model is based on a dual-branch hierarchical encoder-decoder that separately processes spatial and spectral data. Each branch uses local self-attention and cross-attention modules that are fused at multiple stages. Moreover, Ruoppa et al. (2025) introduced a novel unsupervised DL model for the semantic segmentation of forestry MS LiDAR data into foliage and wood components based on GrowSP architecture. To the best of our knowledge, no dedicated 3D DL models for MS LiDAR data have been reported in the literature in LULC classification domain. In our study, we deploy HPF for the first time for MS LiDAR data-based LULC classification. In addition, although Yang et al. (2024) and Zhou et al. (2025) benchmarked some DL models for MS LiDAR-based LULC classification, including PointNet, PointNet++, RandLA-Net, DGCNN, KPConv, PAConv, and PTv1, SOTA DL models for LULC classification are not benchmarked in recent studies. Moreover, since the Optech Titan MS LiDAR system used by Yang et al. (2024) and Zhou et al. (2025) is no longer commercially available, it necessitates exploring the capabilities of currently available MS LiDAR systems. Thus, in this study we benchmark seven SOTA DL models comprising KPConv, KPConvX (Thomas et al., 2024), Superpoint Transformer (SPT; Robert et al., 2023), HPF, PTv1, PTv3 (Wu et al., 2024b), and Sparse UNet (SpUnet; Graham et al., 2018).

To the best of our knowledge, this study represents the first use of KPConvX, SPT, HPF, PTv3, and SpUnet for LULC classification. KPConv is a point-based convolutional operator that processes data using kernel points defined in Euclidean space to flexibly and efficiently capture local geometric features. KPConv has demonstrated promising performance in urban point cloud classification (Bayrak et al., 2023; Bayrak et al., 2024). KPConvX represents an innovative architecture and training strategy of KPConv, enabling scaled depth-wise convolutional weights with kernel attention values. The SPT uses a fast cut pursuit algorithm to hierarchically segment point clouds into superpoints, significantly speeding up the processing (Robert et al., 2023). Its self-attention mechanism further models multi-scale superpoint relationships, enhancing contextual understanding. As stated above, HPF is a novel dual-branch hierarchical encoder-decoder DL model designed to process MS point clouds. PTv1 was the first transformer-based backbone that leverages point-wise vector attention for learning geometric relationships. Building on this foundation, PTv3 also adopts a transformer backbone but emphasizes simplicity, scalability, and efficiency through serialization-based attention to achieve high time-efficiency and accuracy. SpUnet, a UNet-style 3D backbone, employs sparse tensor convolutions to effectively learn multi-scale features by adopting a standard encoder-decoder architecture with residual sparse convolutional blocks and skipping connections to preserve fine-grained spatial details.

Additionally, most studies (13/16) classified no more than seven classes, typically including buildings, trees, grass, roads, soil, powerlines, and water. The maximum number of classes is reported by Reichler et al. (2024), who consider 13 classes. Notably, none of the studies follow a specific standard/scheme for the classes considered. To align LULC classification with the practices and prospective needs of National Mapping and Cadastral Agencies (NMCAs), this study reports on the results of a questionnaire conducted with the support of the European Spatial Data Research (EuroSDR) and European NMCAs. Detailed information about the EuroSDR questionnaire is provided in Section 2.2. We then present the first 3D LULC MS LiDAR benchmark dataset tailored to NMCAs schemes, named Loosdorf-MSL, at two levels of classification detail: current practices of NMCAs (eight classes; LULC-L1) and prospective fine-grained categories (20 detailed classes; LULC-L2) to support the current and future needs of NMCAs and subsequent studies in this domain. The Loosdorf-MSL dataset will be openly available through the TU Wien dataset repository (https://researchdata.tuwien.ac.at/) upon the paper acceptance. The efficiency of the seven mentioned DL models is benchmarked for LULC classification at these two levels of detail. Furthermore, unlike existing studies, we investigate the effects of MS LiDAR data for current and future needs of NMCAs by conducting rigorous spectral ablation studies at both levels. In summary, the main contributions of the work are as follows:

1. Realization and reporting European NMCAs' current and prospective LULC classification schemes using ALS data.
2. Presenting the first 3D LULC MS LiDAR dataset called Loosdorf-MSL, at two levels of detail (eight and 20 classes), based on the current and prospective LULC classification schemes of European NMCAs.

3. Benchmarking seven SOTA DL models for LULC classification, including one specifically designed for MS point cloud classification.
4. Investigating the merits of MS LiDAR data for NMCAs current practices and prospective LULC classification needs by conducting rigorous spectral ablation studies.

This paper is organized as follows: Section 2 provides an overview of LULC classification standards and schemes, including existing frameworks and the EuroSDR questionnaire report. Section 3 introduces the Loosdorf-MSL dataset, describing its characteristics, levels of detail for LULC mapping, its necessity, and a comparison with related benchmark datasets. Section 4 details the proposed methodology for LULC classification, covering preprocessing steps, the DL models used for benchmarking, and the evaluation metrics. Section 5 presents and analyzes the experimental results, including a comparison of DL models' performance and spectral ablation studies. Section 6 discusses the challenges of MS LiDAR data in terms of spectral variability and vegetation class separability. Finally, Section 7 concludes the paper by summarizing key findings and proposing research directions for future studies.

# 2. LULC classification standards and schemes

## 2.1. Existing LULC classification standards and schemes

Significant differences in the number of classes and inconsistent definition of classes are one of the primary factors leading to inconsistency and uncertainty in LULC mapping products (Wang et al., 2023; Walicka and Pfeifer, 2025). Regional and global LULC products mostly follow classification systems developed by the Food and Agriculture Organization of the United Nations (FAO). While these FAO-based LULC maps provide detailed information mainly for forests, other land categories are described more generally (Wang et al., 2023). LADA (Land Degradation Assessment in Drylands) is FAO's most refined LULC classification system encompassing 40 classes (Nachtergaele et al., 2008). This classification system includes several types of forests, grasslands, shrub cover, crops, wetlands, bare areas, sparse areas and water and is specially designed for global land cover mapping scale. The LADA-based LULC map is produced by low-resolution data fusion.

In ALS data classification, the American Society for Photogrammetry and Remote Sensing (ASPRS) has established a predefined classification system that has been used in LAS versions 1.1 (2014)-1.5 (2025). The ASPRS class definitions are widely considered as a standard for ALS-based LULC classification. The latest ASPRS LiDAR class definition includes 16 practical classes (ground, low vegetation, medium vegetation, high vegetation, building, noise, water, rail, road surface, wire-guard, wire-conductor, transmission tower, wire-structure connector, bridge deck, overhead structure, and snow). However, these classes are not fully defined to meet the specific needs of NMCAs for LULC classification.

Recently, Walicka and Pfeifer (2025) proposed a common classification scheme for ALS data that applies to the semantic segmentation of various ALS datasets. By analyzing urban ALS benchmark datasets and additional datasets from Europe, they introduced a unified classification scheme with four classes (ground and water, vegetation, buildings and bridges, and "other") and trained a classifier using ALS data from Austria, Switzerland, and Poland. This approach achieved high accuracy on unseen datasets, demonstrating that unified schemes and joint training effectively boost generalizability across heterogeneous ALS data.

## 2.2. EuroSDR questionnaire on LULC classification schemes

To address the lack of a standardized ALS-based LULC classification tailored to NMCA needs, we conducted a questionnaire, supported by EuroSDR, to identify the classes of interest for European NMCAs in their ALS-based LULC classification schemes. Responses are received from 16 NMCAs across 14 European countries (Netherlands, Denmark, Switzerland, Finland, Sweden, Austria, Germany, Spain, Portugal, Poland, Croatia, Romania, Poland, and Slovenia). The questionnaire results are presented in Fig. 1, where the LULC classes are categorized as a current, planned, or wished classes.

In total, 33 classes are outlined by the NMCAs. Among these, 14 classes are currently used in national point cloud classifications: ground, water, building, wall, low vegetation, medium vegetation, high vegetation, bridge, low/high noise, sports area, ship, snow, jetty, and "other". It is worth noting that, from this point onward, only classes that received responses from at least 10% of NMCAs are considered as current practices, including nine classes. 15 additional finer classes are planned for future implementation, including asphalt, road marking, rocky area, traffic sign, soil, fence, road, vehicle, solar panel, pole, chimney, façade, electric tower, roof, and cable. Furthermore, four classes (i.e., hay bale, window, container, and windmill) are identified as wish classes, indicating that they are not included in the predefined questionnaire options but are suggested as desired additions. Notably, the questionnaire results indicate that ground objects (e.g., road, asphalt, gravel, etc.) are still predominantly classified under a general category (i.e., the ground class). Conversely, vegetation is consistently categorized into three types (i.e. low, medium, and high vegetation) by all participant NMCAs. The questionnaire results reveal that the classification schemes of NMCAs aim toward fine-grained LULC classification.

Considering the reviewed literature on MS LiDAR-based LULC classification (see Table 1) and the summarized report of NMCA's current and prospective LULC classification schemes shown in Fig. 1, it can be substantiated that, while the number of classes considered by Wang et al. (2024) and Wang et al. (2025a) aligns with the current NMCA schemes, there remains a noticeable research gap in exploring the capability of MS LiDAR to address the prospective NMCA schemes (see Fig. 2). Consequently, in this study, we utilize MS LiDAR data for the first time to examine its potential for both current and prospective NMCA LULC classification schemes.

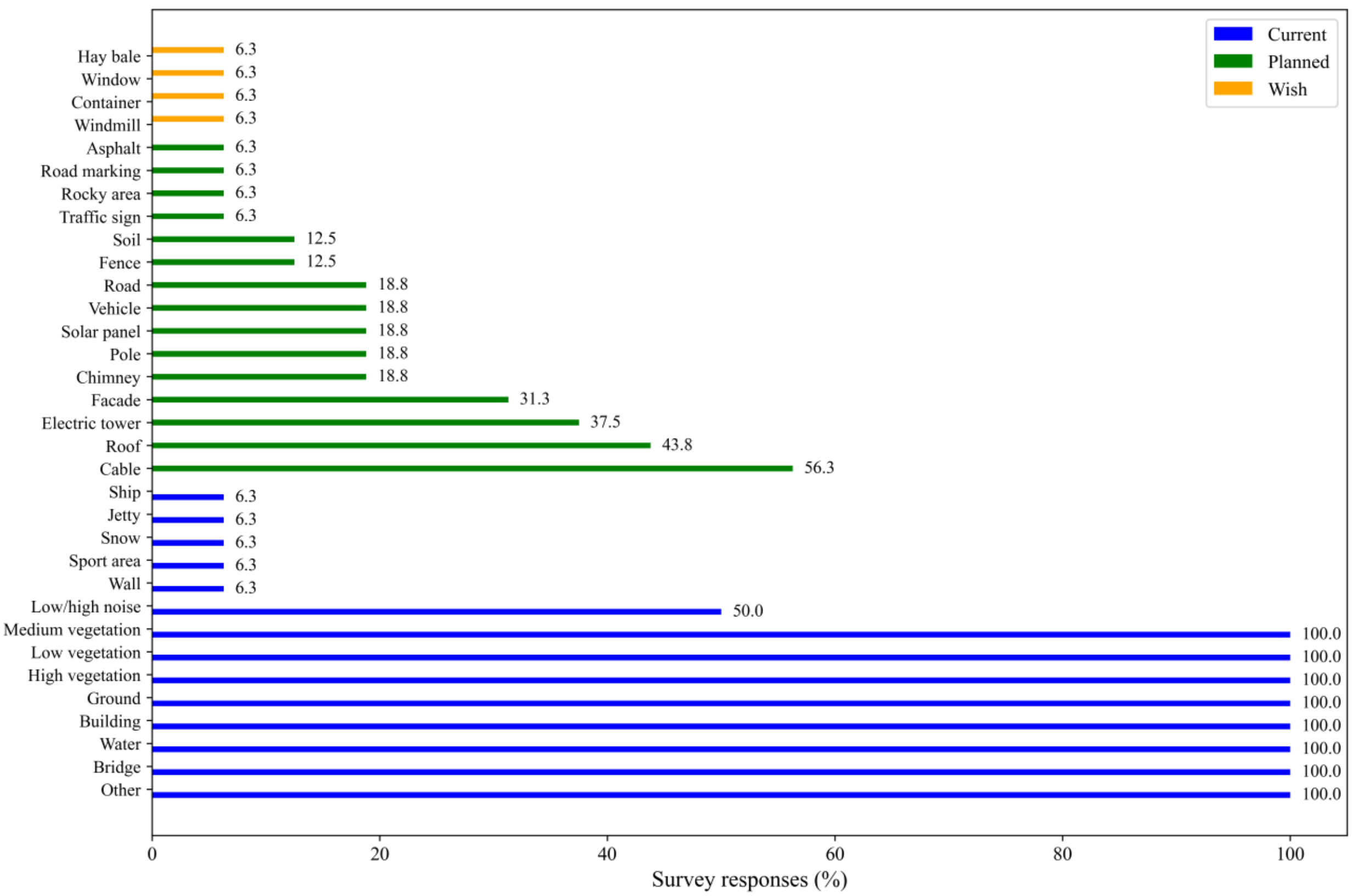

**Fig. 1.** EuroSDR questionnaire on NMCAs' ALS-based LULC classification schemes.

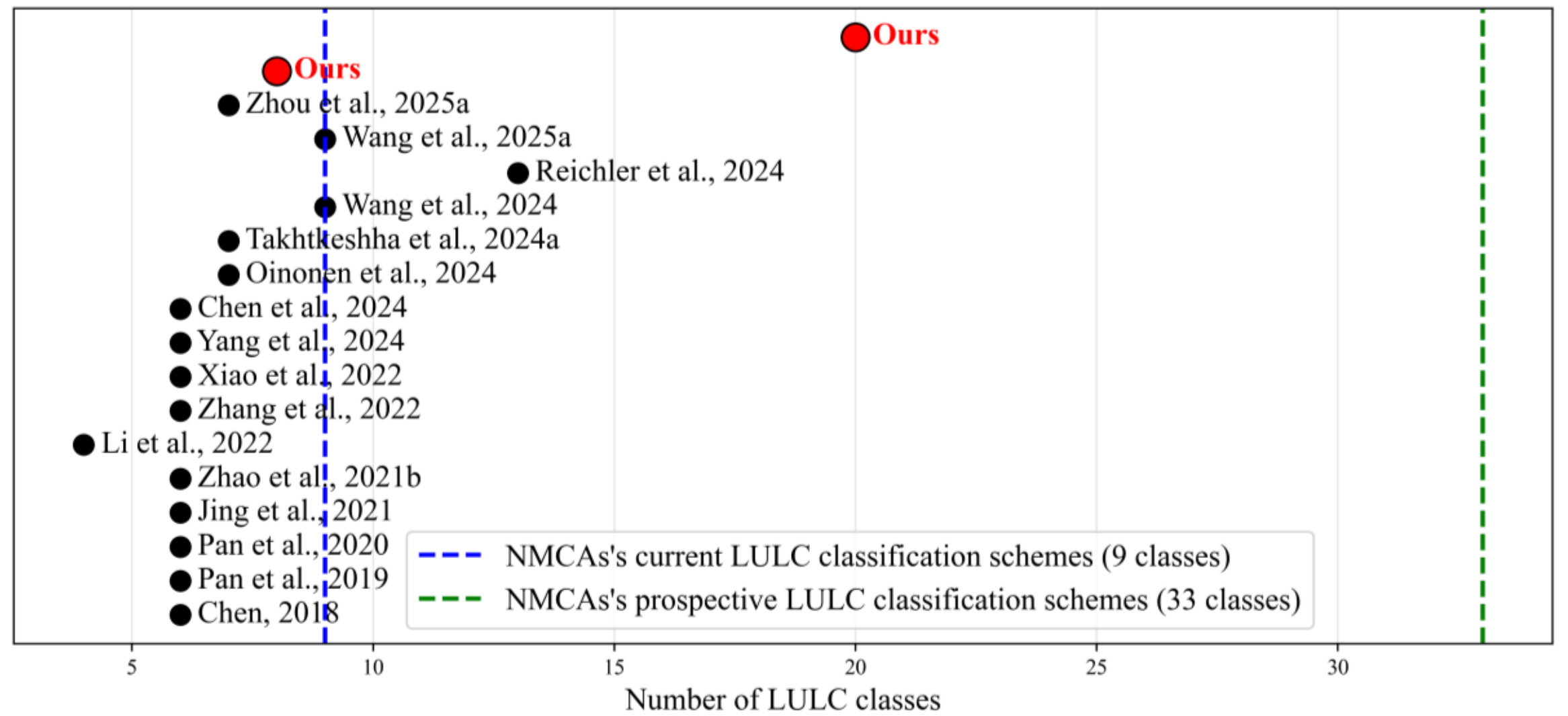

**Fig. 2.** The number of classes considered in the SOTA MS-LiDRA and DL-based studies vs NMCAs LULC classification schemes.

## 3. Material: Loosdorf-MSL dataset

The Loosdorf-MSL dataset was acquired in October 2023 using a commercial MS ALS system, *RIEGL* VQ-1560i-DW, operating at green (532 nm) and NIR (1064 nm) wavelengths over the village of Loosdorf and the city of Melk in Lower Austria (Fig. 3a). Table 2 summarizes the specifications of the Loosdorf-MSL dataset. The average point densities for the green and NIR scanners are 9.1 points/m² and 14.7 points/m², respectively, while the combined MS point cloud has an average density of 47.1 points/m². The acquired laser data and nadir images are precisely georeferenced and co-aligned through a hybrid adjustment process using the StripAdjust[1] module in OPALS software (Mandlburger et al., 2009; Pfeifer et al., 2014). This module enhances the georeferencing of ALS data and aerial imagery by rigorously combining strip adjustment and aerial triangulation (Glira et al., 2019). Photogrammetric point clouds and orthophotos (at 20 cm Ground Sampling Distance, GSD) are generated by dense image matching using SURE nFrames (Esri, version 2024.3.4) and are provided in the Loosdorf-MSL dataset as auxiliary data. The validation of the hybrid adjustment is reported in Appendix A.
The Loosdorf-MSL dataset covers 1.6 km² of suburban and forested landscapes and comprises 103,238,318 manually labeled points categorized into 8 and 20 classes, which are used for training and evaluating DL models. Point clouds are manually annotated using CloudCompare software (Girardeau-Montaut, 2021), with support of auxiliary photogrammetric point clouds and field survey data. LULC classes are defined based on the current and prospective ALS-based LULC classification schemes of NMCAs (Section 2.2). More information about these two levels of detail is provided in Section 3.1. By considering the recent TerLiDAR (Carós et al., 2026) ALS benchmark dataset, the Loosdorf-MSL dataset is divided into training, testing, and validation plots in a 71:17:12 ratio. The dataset includes a total of 15 plots, six of which are used for training and four for testing (Fig. 3b).

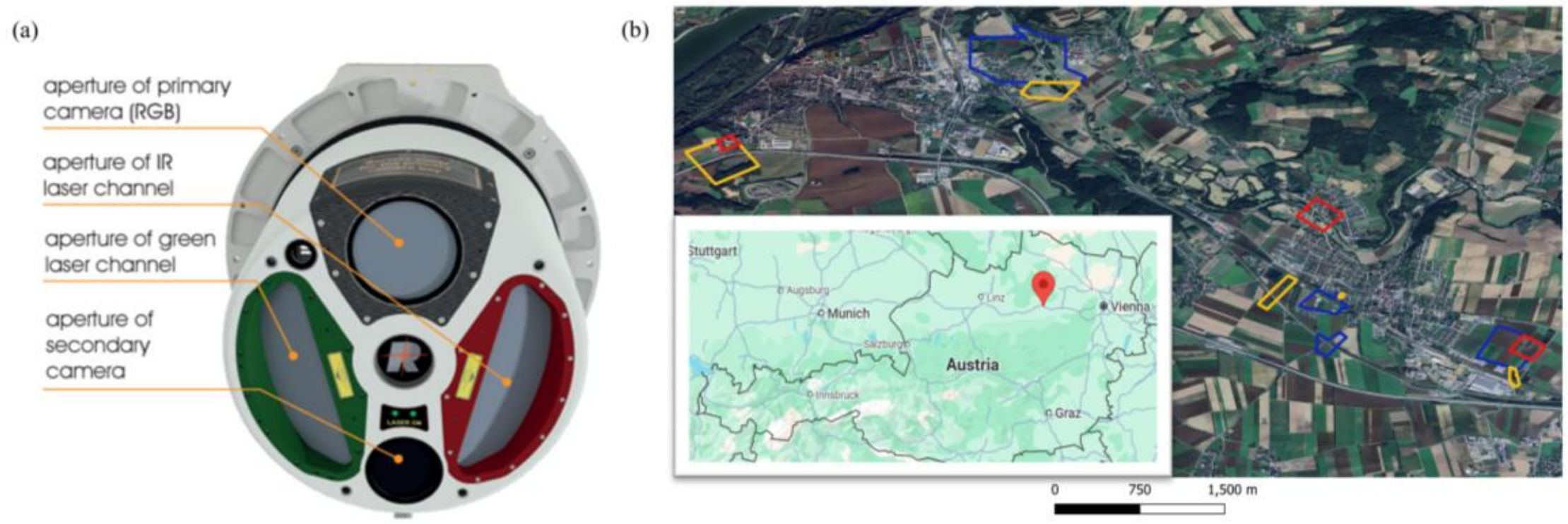

**Fig. 3.** (a) *RIEGL* VQ-1560i-DW laser scanner; (b) Loosdorf-MSL dataset plots, where training plots are shown in blue, validation plots in yellow, and test plots in red.

**Table 2** Specifications of the Loosdorf-MSL dataset.

| MS LiDAR system | Wavelength (nm) | Point density (point/m2) | Pulse repetition rate (kHz) | Laser beam divergence (mrad) | Flight altitude (m) |
|---|---|---|---|---|---|
| VQ-1560i-DW | 532 | 9.1 | 1000 | 2.2 | 700m |
| | 1064 | 14.7 | 1000 | 0.3 | |

### 3.1. Levels of detail for LULC classification

To evaluate the effectiveness of MS LiDAR data for both current and prospective fine-grained LULC classification applications tailored to NMCAs schemes, the Loosdorf-MSL dataset is prepared at two levels of detail as summarized in Table 3. The first classification level (LULC-L1) reflects the current requirements of NMCAs and includes eight classes: ground, water, low vegetation, medium vegetation, high vegetation, building, bridge, and “other”.
The vegetation categories are defined as follows:

- Low vegetation: vegetation with 0.3 m ≤ height above ground < 2 m
- Medium vegetation: vegetation with 2 m ≤ height above ground < 5 m
- High vegetation: vegetation with height above ground ≥ 5 m

The second level (LULC-L2) incorporates all classes currently used, planned, or desired by NMCAs. Fig. 4 illustrates the transition from LULC-L1 (current NMCA schemes) to LULC-L2 (prospective schemes). For instance, the general ground class in LULC-L1 is subdivided into asphalt, soil, roads, road markings, and sports areas in LULC-L2. Similarly, electricity infrastructure, which falls under the "other" class in LULC-L1, is further divided into electric towers, cables, and poles in LULC-L2. Consequently, a total of 20 classes is included in LULC-L2, representing perspective and fine-grained LULC classification.
It is important to note that at both levels of detail, some underrepresented classes are merged into broader categories. In effect, jetty is included under the bridge class. Furthermore, there are a few gravel areas in our study areas which are included under the asphalt class. Additionally, class "other" in L2 generally contains miscellaneous urban elements, including trash bins, containers, and other items.

[1] https://opals.geo.tuwien.ac.at/html/stable/ModuleStripAdjust.html

**Table 3** Considered levels of detail for LULC classification based on the conducted EuroSDR questionnaire.

| Level of detail | Description | Classes |
|---|---|---|
| LULC-L1 | According to the current European NMCAs' practices | 8 (ground, water, low vegetation, medium vegetation, high vegetation, building, bridge, and "other") |
| LULC-L2 | Fine-grained LULC classification encompassing all observed classes in our case study including current, planned, and wish classes pointed out by European NMCAs. | 20 (asphalt, soil, road, water, low vegetation, medium vegetation, high vegetation, roof, façade, chimney/roof objects, solar panel, vehicle (car and truck), electric tower, cable, pole, bridge, fence/wall, sport area, road marking, and "other") |

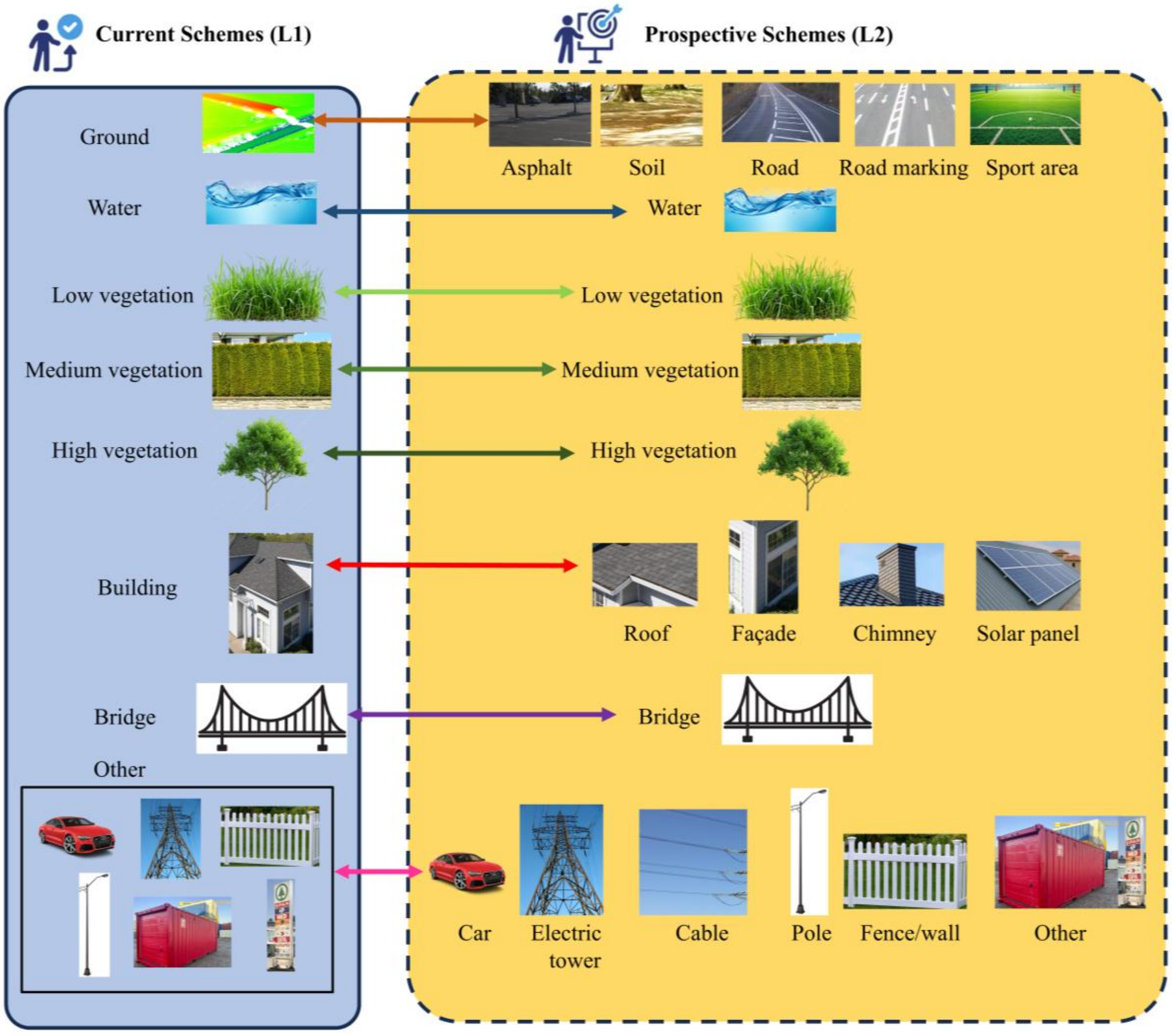

**Fig. 4.** Transition from LULC-L1 (current schemes of NMCAs) to LULC-L2 (prospective schemes).

An example plot (train 1) of the Loosdorf-MSL dataset is shown in Fig. 5, presenting the orthophoto, NIR reflectance, green reflectance, and ground truths at L1 and L2. Fig. 6 and Fig. 7 depict the class distribution at the two levels of detail. As shown, the heterogeneous suburban/urban Loosdorf-MSL dataset is highly imbalanced at both levels, making the LULC classification more challenging. Also, Fig. 8 illustrates the spectral histograms of various LULC classes, highlighting the unique MS signatures characteristic of each object. These distinct spectral behaviors underscore the potential of MS LiDAR data to enhance the accuracy of LULC classification. In this study reflectance values, which are range-corrected intensities calibrated against a white reference target are considered as spectral information. *RIEGL*'s V-Line laser scanners provide such reflectance values (Pfennigbauer and Ullrich, 2010). For visualization purposes, reflectance values are normalized between 0 and 1.

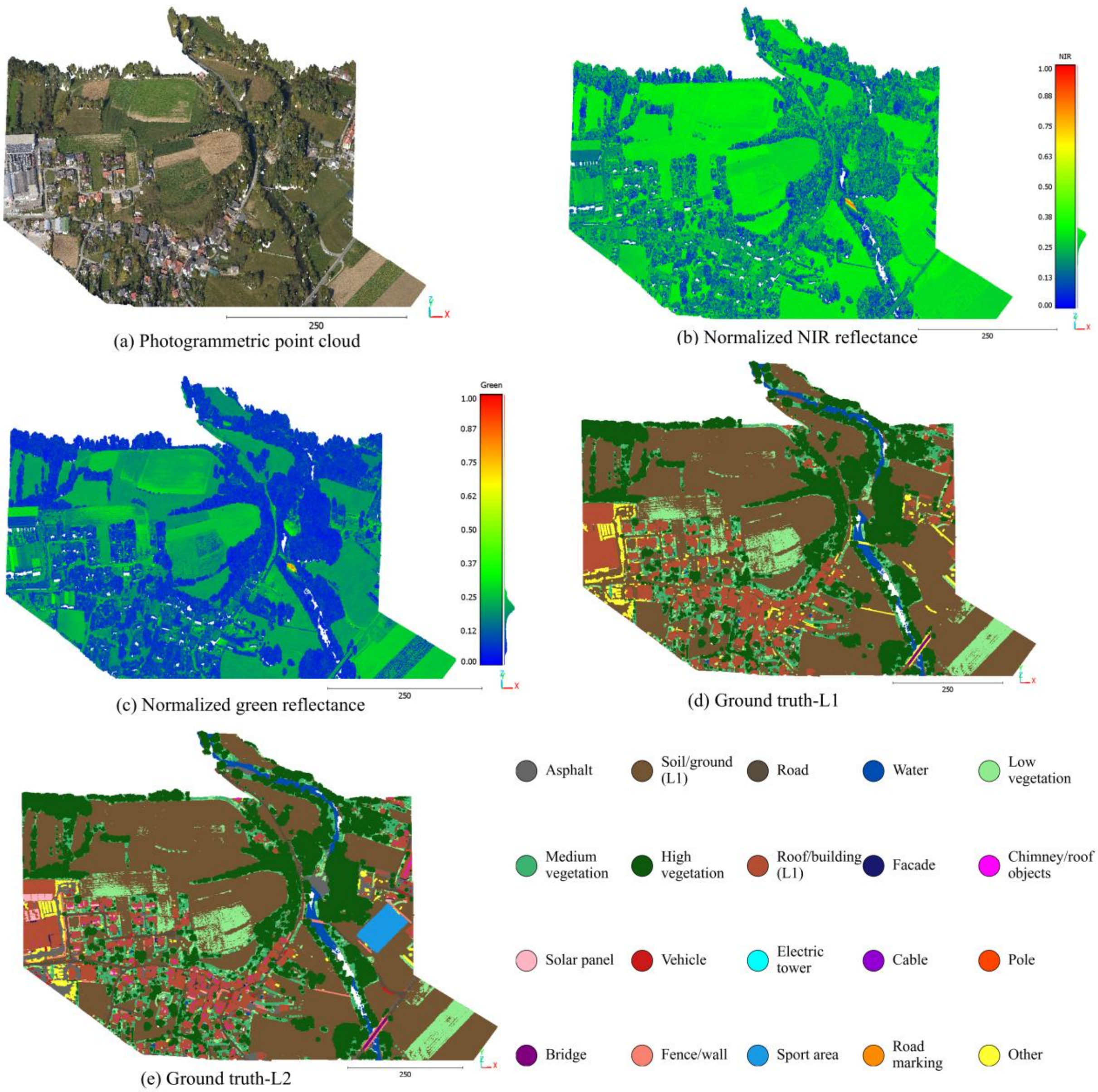

**Fig. 5.** Example of the Loosdorf-MSL benchmark dataset (train 1 plot): (a) photogrammetric point cloud, (b) normalized NIR reflectance, (c) normalized green reflectance, (d) ground truth L1, (e) ground truth $L^2$. Reflectance values are normalized to the range [0, 1] for visualization.

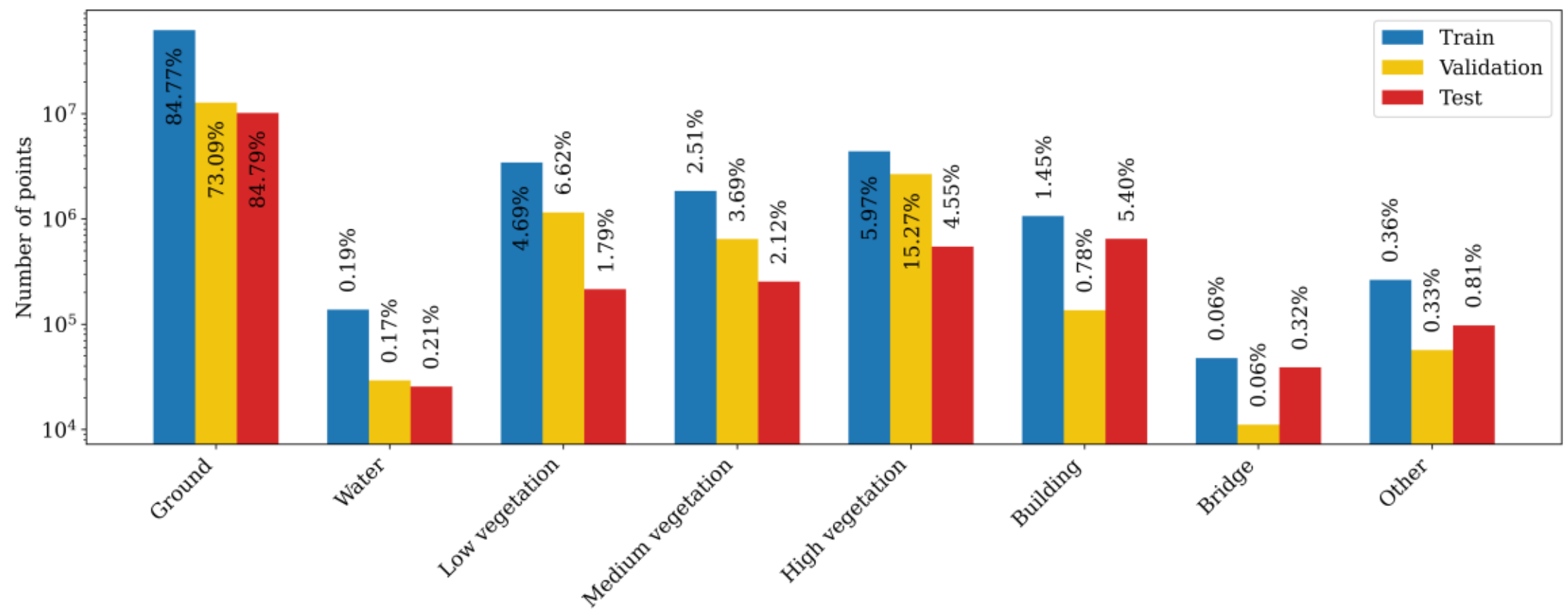

**Fig. 6.** Class distribution within the Loosdorf-MSL dataset at LULC-L1.

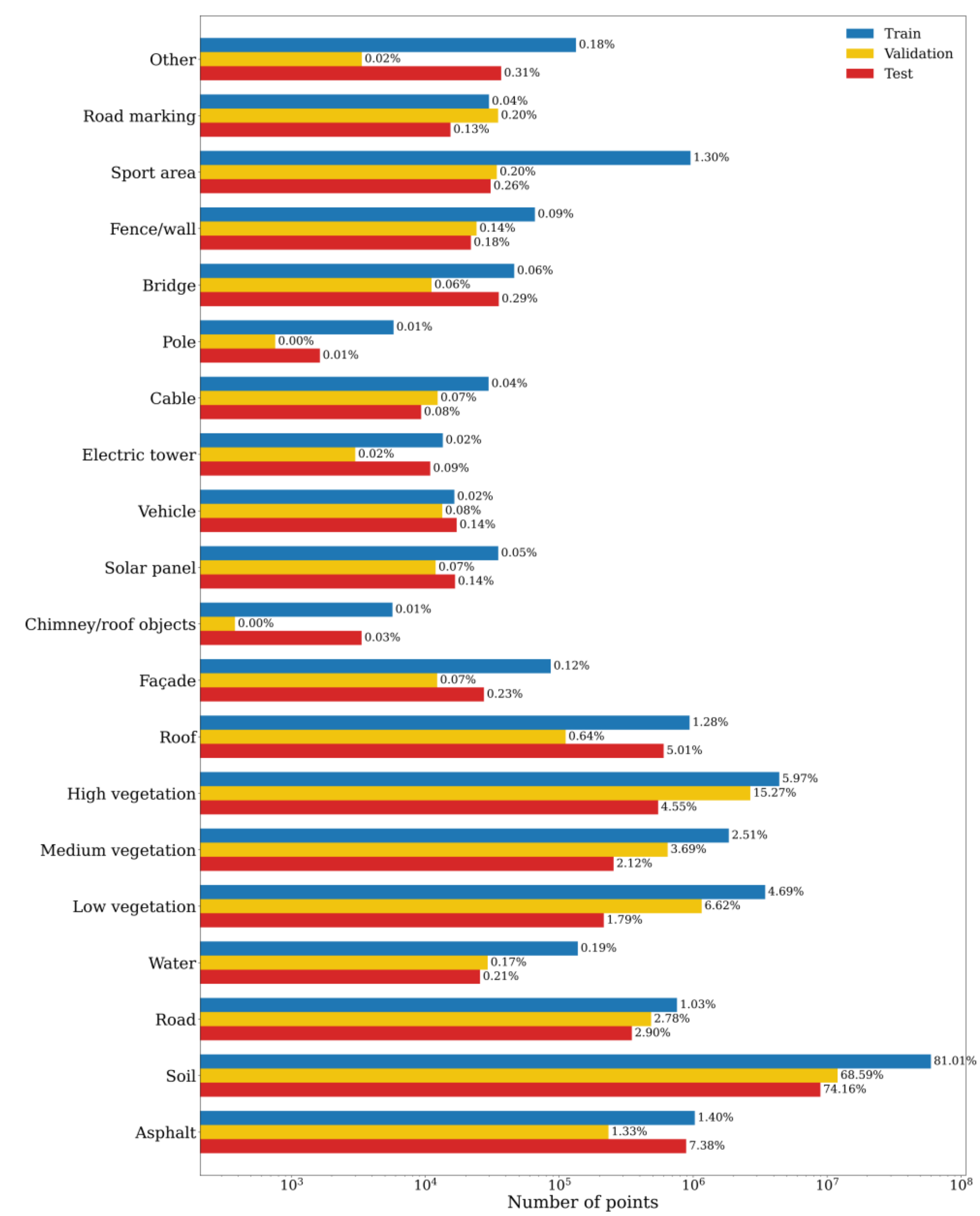

**Fig. 7.** Distribution of classes in the Loosdorf-MSL dataset at LULC-L2.

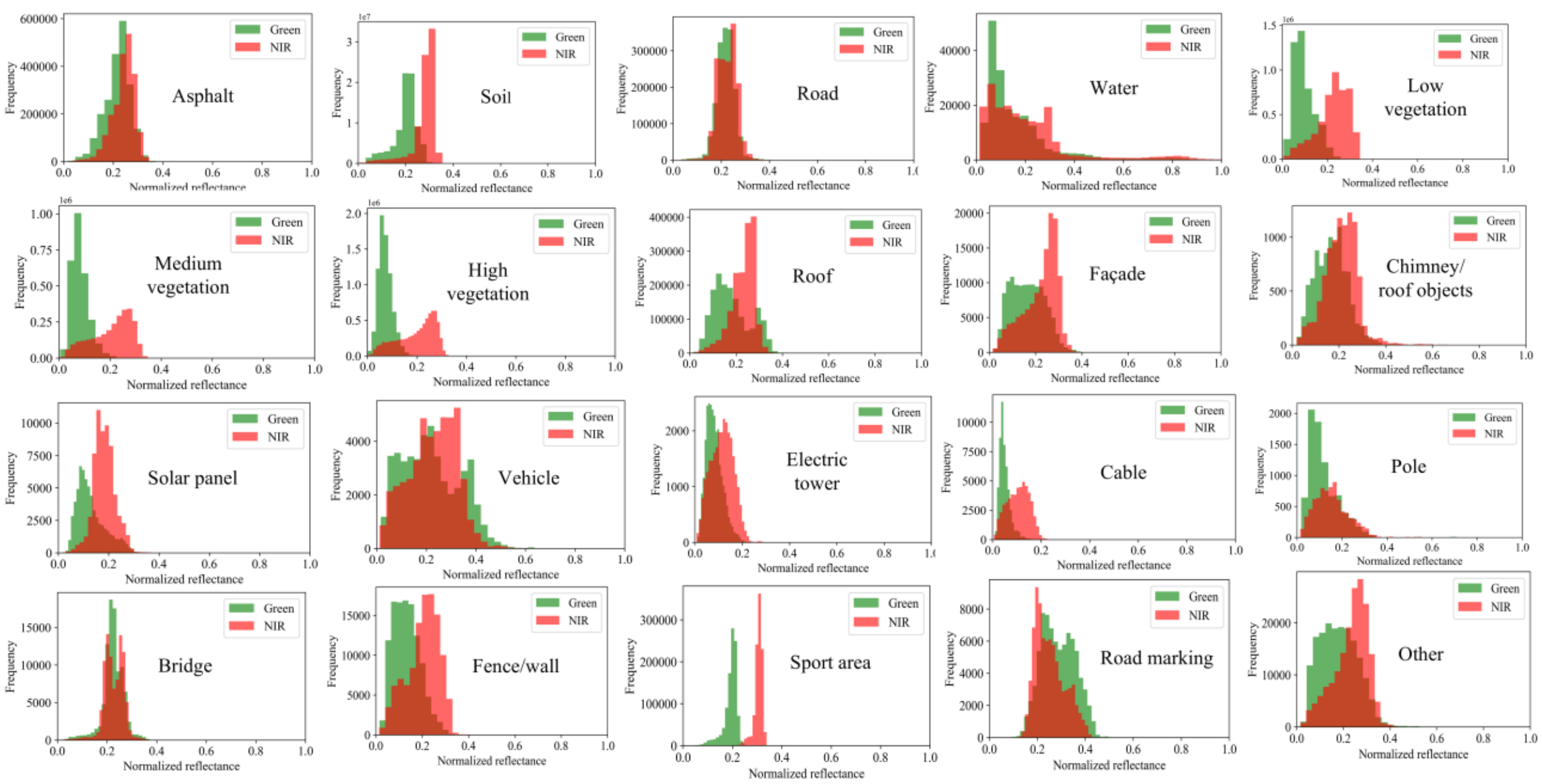

**Fig. 8.** Spectral histogram of different LULC classes across different laser wavelengths. Reflectance values are scaled between 0 and 1 for visualization purposes.

### 3.2. Importance of the Loosdorf-MSL dataset and comparison with existing datasets

To date, considerable efforts exist in preparing ALS/ULS (UAV-LiDAR)-based LULC classification benchmark datasets, specifically to facilitate the training, evaluation, and development of data-hungry DL models, beginning with the ISPRS dataset (Niemeyer et al., 2014; 9 classes). Among these benchmarks, some contain a high number of classes (more than 10): DublinCity (Zolanvari et al., 2019; 13 classes), Hessigheim 3D (Kölle et al., 2021; 11 classes), WHU3D (Han et al., 2024; 37 classes), and TerLiDAR (11 classes).

**Table 4** Comparison of the Loosdorf-MSL benchmark dataset with other suburban/urban MS LiDAR benchmark datasets. MS LiDAR datasets are marked with *.

| Benchmark dataset | Sensors | Number of classes | Spatial size ($km^2$) | Point density (points/$m^2$) | Labeled points (millions) | Channels | LULC scheme | Auxiliary data |
|---|---|---|---|---|---|---|---|---|
| *DFC2018 | Optech Titan, ALS | 20 | 5 | 15 | NA | 1550 nm, 1064 nm, and 532 nm | × | RGB and hyperspectral orthophotos |
| DublinCity | ALS | 13 | 2 | 250-348 | 260 | 1064 nm | × | NA |
| Hessigheim | ULS | 11 | 0.08 | 800 | 125 | 1064 nm | × | photogrammetric point clouds (RGB) |
| WHU3D | ALS + MLS | 37 | 0.0065 | 35 | 393 | 1064 nm | × | NA |
| TerLiDAR | ALS | 11 | 51.4 | 8-16 | 692 | 1064 nm | × | photogrammetric point clouds (RGBI) |
| *FGI-EMIT | HeliALS-TW, ALS | 5 | 0.04 | 1660 | 60 | 1550 nm, 905 nm, and 532 nm | × | NA |
| *Loosdorf-MSL (ours) | VQ-1560i-DW, ALS | 8/20 | 1.7 | 47.1 | 103 | 1064 nm and 532 nm | NMC As | Hybrid adjusted photogrammetric point clouds and RGB orthophotos |

A summary comparing the Loosdorf-MSL with other ALS/ULS benchmark datasets for LULC classification is presented in Table 4. Notably, none of these LULC benchmarks are MS LiDAR data. Furthermore, they do not follow any specific scheme/standard in class definition. Another merit of the Loosdorf-MSL over existing multi-source point cloud benchmark datasets (like TerLiDAR) is that the aerial images and LiDAR data are precisely georeferenced and co-aligned through a hybrid adjustment process. Moreover, some datasets, such as Hessigheim, are annotated in 2.5D rather than fully 3D, containing a mixture of low vegetation and the underlying soil within the low vegetation class. This limitation reduces the usability of such datasets for NMCAs schemes, in which the ground class is imperative for DTM generation.

Based on the available literature, only two MS LiDAR benchmark datasets for LULC classification are currently available. The first benchmark dataset is associated with the IEEE Geoscience and Remote Sensing Society (GRSS) data fusion contest in 2018 (DFC2018 dataset, Prasad et al., 2020). This dataset is prepared by the University of Houston and covers its campus and surrounding neighborhood (up to 5 km$^2$). It was acquired by the National Center for Airborne Laser Mapping (NCALM) on 16 February 2017. The dataset comprises MS point clouds collected by the three-wavelength Optech Titan MS LiDAR system, RGB images at 5 cm GSD, hyperspectral images (48 bands ranging from 380 nm to 1050 nm with 1 m GSD), and rasterized ground truth (5 cm GSD). The DFC2018 dataset comprises 20 classes, including healthy grass, stressed grass, artificial turf, evergreen trees, deciduous trees, bare earth, water, residential buildings, non-residential buildings, roads, sidewalks, crosswalks, major thoroughfares, highways, railways, paved parking lots, unpaved parking lots, cars, trains, and stadium seats. Although the DFC2018 dataset has continued to attract attention within the research community and has been utilized in recent studies (Zhou et al., 2025a; Rizaldy et al., 2025), its main drawbacks lie in the 2D nature of the ground truth and the sparsity of labels, as most pixels remain unlabeled (see Fig. 9). The 2D annotation of DFC2018 significantly hampers its usability for 3D LULC classification. In addition, like other available datasets, the DFC2018 dataset does not adhere to any established LULC classification scheme/standard.

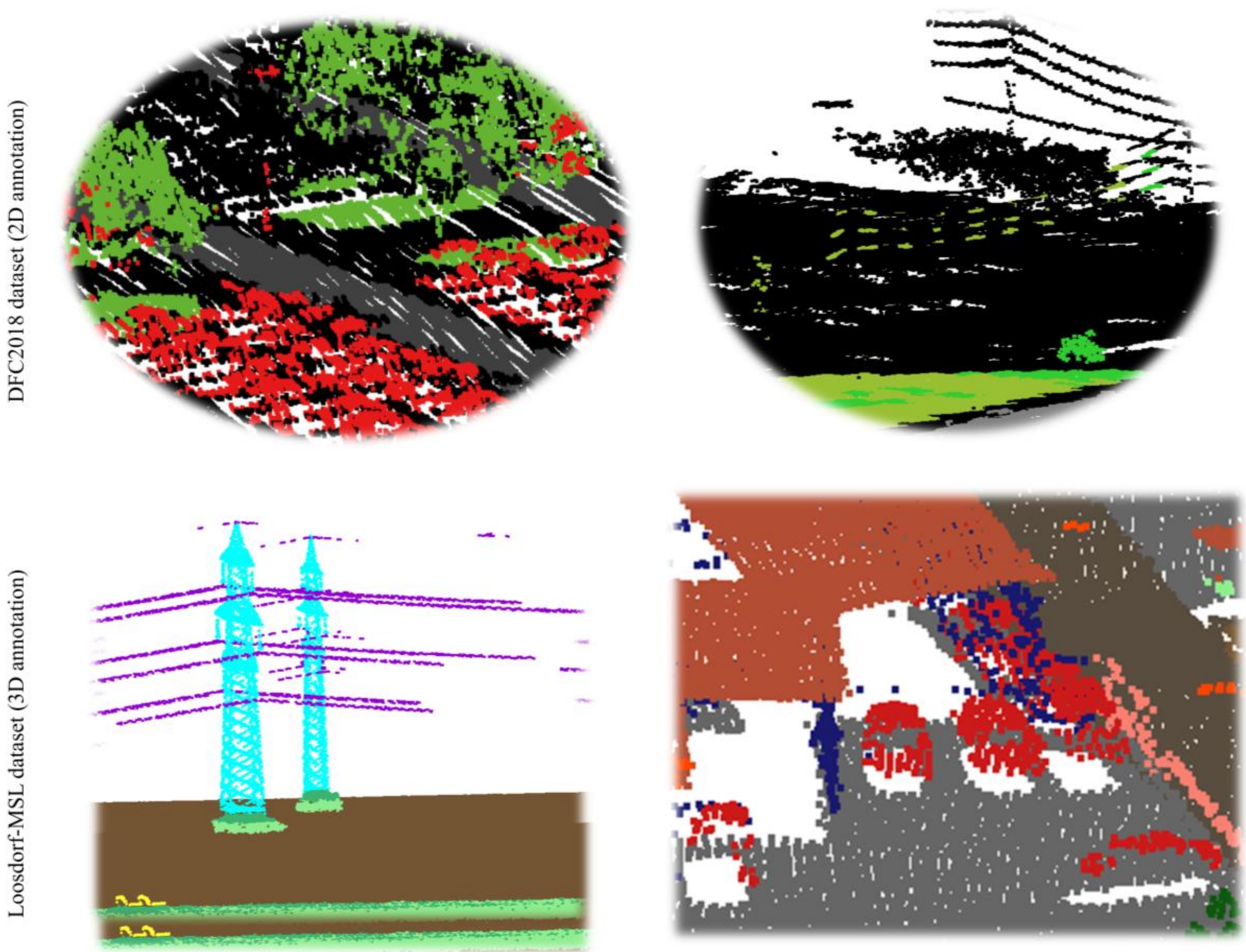

**Fig. 9.** 2D annotated point clouds in the DFC2018 dataset (a) vs 3D annotated point clouds in the Loosdorf-MSL dataset (b). The black points in the DFC2018 dataset denote unannotated points.

Recently, Ruoppa et al. (2026) presented a new benchmark dataset called FGI-EMIT. This dataset is tailored for individual tree segmentation using three-wavelength high-density MS ALS data (HeliALS-TW). Although FGI-EMIT is specifically introduced for individual tree segmentation, it also provides ground truth annotations for classes other than trees. In fact, the FGI-EMIT dataset includes five classes: tree, building, vehicle, pole, and "other". Nevertheless, FGI-EMIT has not yet been used in any study for LULC classification purposes. More importantly, most plots in this dataset represent pure forestry areas, while only a few plots contain urban classes as well. Another limitation of the FGI-EMIT dataset is that the classes are not as distinct as their names suggest. For example, the building class also contains structures such as fences and sheds, while the pole class includes traffic signs. Understory vegetation and ground points are grouped under the "other" class, together with several additional categories. Notably, similar to DFC2018, FGI-EMIT does not follow any standardized classification scheme. Furthermore, the coverage extent of FGI-EMIT is considerably lower than that of the DFC2018 and the Loosdorf dataset and does not contain any auxiliary data. The proposed Loosdorf-MSL represents the first publicly available 3D MS LiDAR benchmark dataset aligned with NMCAs' LULC classification schemes. In addition to being MS LiDAR data, another distinctive merit of this dataset is that it is prepared in accordance with the current practices and prospective needs of NMCAs. It also provides detailed building elements, including solar panels, chimneys, roofs, and facades. Additionally, the Loosdorf-MSL dataset exhibits a substantially higher point density compared to DFC2018 dataset. Moreover, unlike DFC2018 dataset, the auxiliary photogrammetric point clouds and RGB orthophotos are co-aligned with LiDAR data through a hybrid adjustment. However, compared with the DFC2018 and FGI-EMIT datasets, Loosdorf-MSL dataset includes two wavelengths instead of three. Also, compared to the DFC2018 dataset, our dataset has a smaller spatial coverage and does not include hyperspectral imagery.

## 4. Methodology for LULC classification

The overall workflow of the proposed methodology for MS LiDAR-based LULC classification is shown in Fig. 10, comprising three main steps:

1. Data processing and preparation of the Loosdorf-MSL benchmark datasetBenchmarking of different SOTA 3D DL models.
2. Effectiveness assessment of MS LiDAR-based LULC classification through rigorous spectral ablation studies.

The first two steps are described in Sections 4.1–4.2, while the spectral ablation study is detailed in Section 5.2.

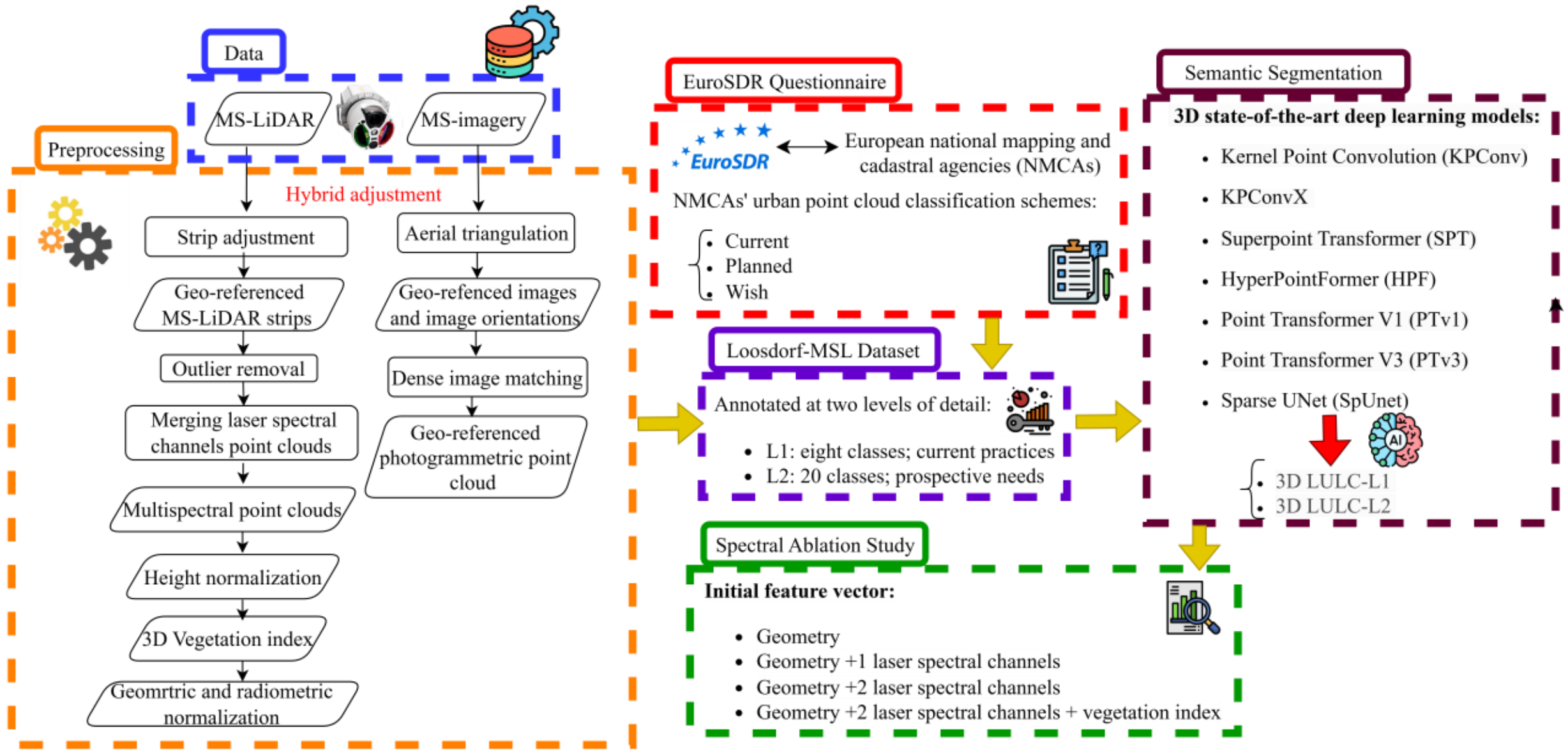

**Fig.10.** The proposed pipeline for MS LiDAR-based LULC classification.

### 4.1. Preprocessing

After performing hybrid adjustment to georeference and co-align MS LiDAR and imagery data as described in Section 3, point clouds from each channel are denoised using a Statistical Outlier Removal (SOR) filter in CloudCompare prior to merging laser spectral channels point clouds. Subsequently, MS point clouds are generated by merging the point clouds from the two channels using the preMergeChannelsPointclouds[2] script in OPALS. During the merging of point clouds from individual channels to maintain the reliability of the spectral information, only the nearest neighboring point is considered when assigning spectral information to each point in the MS point cloud from other channels, Also, a search radius of 7 m in 3D space is applied to preserve points of façade and other large objects. Reflectance values of both channels are considered for MS point cloud generation.

[2] https://opals.geo.tuwien.ac.at/html/stable/preMergeChannelsPointclouds.html

$$\text{Reflectance}_{\text{linear}} = 10^{\frac{\text{Reflectance}_{\text{db}}}{10}} \quad (1)$$

$$\text{pNDVI} = \frac{(\text{NIR}_{\text{linear}} - \text{green}_{\text{linear}})}{(\text{NIR}_{\text{linear}} + \text{green}_{\text{linear}})} \quad (2)$$

$$\hat{f} = \frac{f - \text{median}(f)}{\text{IQR}(f)} \quad (3)$$

$$\text{f_normalized} = \frac{\hat{f} - \min(\hat{f})}{\max(\hat{f}) - \min(\hat{f})}$$

The point cloud heights are normalized by first identifying ground points using the Cloth Simulation Filter (CSF) plugin in CloudCompare, followed by generating a Digital Terrain Model (DTM) from these ground points. The normalized height values (i.e., height above ground) are then computed by subtracting each point's height from the local minimum height using OPALS AddInfo module.

Using the NIR and green channels, the Pseudo Normalized Difference Vegetation Index (pNDVI) is calculated from the reflectance values in linear units (Eq. 1 and 2) and is employed as an additional spectral feature for the DL models (only in spectral ablation studies). Finally, the MS point clouds undergo both geometric and radiometric normalization. In terms of geometry, the clouds are recentered in the planimetric axes (X and Y) by subtracting their mean planimetric coordinates from the planimetric coordinates. The heights of point clouds are shifted by subtracting the minimum height so that all heights become positive. For radiometric normalization (Eq. 3), outlier-robust scaling ($\hat{f}$) is combined with min–max scaling (f_normalized), where f represents the spectral attribute and the Interquartile Range (IQR) corresponds to the difference between the 75th and 25th percentiles.

## 4.2. DL models used for benchmarking

To evaluate the efficiency of DL models for MS LiDAR-based LULC classification, seven SOTA 3D DL models are benchmarked: KPConv (Thomas et al., 2019), KPConvX (Thomas et al., 2024), SPT (Robert et al., 2023), HPF (Rizaldy et al., 2025), PTv1 (Zhao et al., 2021a), PTv3 (Wu et al., 2024b), and SpUnet (Graham et al., 2018).

**Table 5** The main hyperparameters of the benchmarked DL models.

| Model | Grid/voxel size (m) | Batch size | Initial LR | Weight decay | Number of epochs | Additional parameters |
|---|---|---|---|---|---|---|
| KPConv | 0.1 | 6 | 0.01 | 0.05 | 300 | Input sphere radius= 20 m |
| KPConv X | 0.1 | 6 | 0.0001 | 0.01 | 300 | Input sphere radius= 15 m |
| SPT | 0.1 | 1 | 0.01 | 0.05 | 1,000 | Number of nearest neighbors= 15; maximum search radius= 40 m (L1) and 10 (L2); regularization= [0.1, 0.2, 0.9] at L1 and [0.1, 0.1, 0.1] at L2; spatial_weight= [0.1, 0.01, 0.001]; cutoff= [10, 40, 80] at L1 and [10, 20, 80] at L2 |
| HPF | 0.1 | 16 | 0.001 | 0.05 | 1,000 | Initial cross-attention weight = 0.01, OneCycleLR scheduler |
| PTv1 | 0.1 | 16 | 0.006 | 0.05 | 1,000 | MultiStepLR scheduler |
| PTv3 | 0.1 | 16 | 0.001 | 0.05 | 1,000 | Serialization patterns: Z-order, Trans z-order, Hilbert, and Trans Hilbert |
| SpUnet | 0.1 | 16 | 0.002 | 0.005 | 1,000 | PolynomialLRD scheduler |

Given that the considerable superiority of recent SOTA DL models over PointNet, PointNet++, DGCNN, and RandLA-Net in LULC classification has already been demonstrated by Yang et al. (2024) and Zhou et al. (2025), these models are excluded from our benchmarking. We use the official KPConvX[3] and SPT[4] repositories to implement KPConv and KPConvX, and SPT, respectively. We also use the official Pointcept[5] repository to implement PTv1, PTv3, and SpUnet. The Pointcept-based implementation of HPF[6] is utilized for HPF. All models are trained first by incorporating spatial information together with the basic spectral information of MS LiDAR data (i.e., reflectance values of all channels). Subsequently, spectral ablation studies are conducted on the best-performing model, considering different spatial and spatial–spectral input feature scenarios. Although all benchmarked DL models except for HPF are not specifically designed for MS point clouds, they are commonly implemented with RGB features, making them flexible in accommodating different input modalities. Therefore, we incorporate the spectral information derived from MS LiDAR data alongside coordinates and normalized height.
From an architectural perspective, all models used in our experiments adopt or are integrated into hierarchical encoder–decoder pipelines like U-Net, except for SPT. The encoder extracts rich features at different scales by gradually down sampling points, then the decoder recovers the details to match the input points. KPConv, KPConvX, and SpUnet use the convolutional operation for learning local features, while HPF, PTv1, and PTv3 use local point aggregation and self-attention for learning. SPT, on the other hand, relies on a hierarchical partitioning strategy that groups geometrically and spectrally homogeneous points into superpoints, along with a self-attention mechanism that captures relationships between superpoints across multiple scales. The key set hyperparameters for each model are summarized in Table 5. Hyperparameters for all models are initially set based on their original implementations for an urban benchmark dataset and subsequently fine-tuned empirically. To ensure a fair comparison among the different DL models, AdamW is set as the optimizer for all the models. The Weighted Cross-Entropy (WCE) loss function is used for KPConv, KPConvX, and SPT, while a combination of WCE and Lovász-Softmax loss functions is applied to the Pointcept-based implemented models (i.e., HPF, PTv1, PTv3, and SpUnet). It is worth noting that, unlike the other models, the Pointcept-based implementations natively support the combined WCE and Lovász-Softmax loss functions. We tested the use of combined loss functions also for KPConv, KPConvX, and SPT; however, this did not yield any improvement. Therefore, we retain the default WCE loss function for these models. In the WCE loss function, class-specific weights are assigned according to their frequencies in the dataset. This loss-based weighting helps the model to better learn underrepresented classes. The grid size of all models is set to 0.1 m for a fair comparison. The default Learning Rate (LR) scheduler used for KPConv and KPConvX is a cyclical LR scheduler with warmup and additional multiplicative decay. For SPT, the default scheduler is cosine annealing with a warmup phase. In contrast to the repositories of KPConv, KPConvX, and SPT, the Pointcept codebase, which is used to implement HPF, PTv1, PTv3, and SpUnet provides greater flexibility in selecting the LR scheduler. Among the available options, after conducting several experiments, the One Cycle LR (OneCycleLR) scheduler is used for HPF and PTv3. Meanwhile, the Multi-step LR (MultiStepLR) scheduler is used for PTv1, and Polynomial LR Decay (PolynomialLRD) is used for SpUnet. This is because our experiments revealed that HPF and PTv3 are highly sensitive to high LR at the beginning of the training phase, often resulting in NaN loss. When using the OneCycleLR scheduler, the session begins with a low LR value, which is then increased to a maximum value and subsequently reduced again in stages. The initial use of a small LR helps prevent NaN loss. In contrast, PTv1 and SpUnet are more robust and can tolerate higher LR in the early stages of training, leading to faster convergence.
Unlike other DL models, SPT intrinsically computes several geometric features, such as verticality and linearity, within the model itself and uses them to generate superpoints and subsequently train the model. Following the DALES configuration, linearity, planarity, scattering, verticality, normal, and normalized height are considered as geometric features. These initial features are utilized for both superpoint generation and segmentation.
It should be noted that fine-tuning all seven SOTA DL models implemented in this study is not a straightforward task, particularly for our new suburban/urban MS point cloud dataset. Nevertheless, we have tried to fine-tune the models fairly and efficiently, considering the limitations of our computing system, by conducting several experiments. All experiments are conducted on two systems. The first one is a Windows system with Windows Subsystem for Linux (WSL), equipped with two Intel® Xeon® Gold 6226R CPUs operating at 2.90 GHz and three NVIDIA RTX A6000 GPU with 48 GB of dedicated memory for each GPU. The second system is a Linux system with 128-core CPU and eight NVIDIA A100 SXM4 Tensor Core GPUs, each providing 80 GB memory.

### 4.3. Evaluation metrics

Results are evaluated using well-established metrics, including Overall Accuracy (OA), Mean Accuracy (mAcc), Intersection over Union (IoU), and mIoU. Among these evaluation metrics, mIoU is commonly prioritized for selecting the best-performing model in LULC point cloud classification (Zhou et al., 2025a; Wang et al., 2025) because it evaluates segmentation quality on a per-class basis, thereby mitigating the bias toward large-area classes inherent in OA. It also provides a balanced measure of spatial overlap between predicted and ground truth classes, which is critical for accurate LULC delineation. Moreover, compared to mAcc, mIoU offers a stricter and more informative evaluation, as it accounts for both false positives and false negatives, penalizes misaligned boundaries, and better reflects the spatial correspondence of predicted and true regions.

## 5. Results

[3] https://github.com/apple/ml-kpconvx
[4] https://github.com/drprojects/superpoint_transformer
[5] https://github.com/Pointcept/Pointcept/tree/df36980119f4636beb2d02d04ef3b2fec0fddfba
[6] https://github.com/aldinorizaldy/hyperpointformer

The results of the different DL models benchmarked for LULC classification at L1 and L2, as well as those of the spectral ablation study, are reported in Sections 5.1 and 5.2, respectively.

### 5.1. Comparison of DL models

The results of MS LiDAR-based LULC classification using different SOTA DL models at levels LULC-L1 and LULC-L2 are presented in Tables 6 and 7, respectively. These tables compare the capability of various DL models to process MS LiDAR data, using an initial feature vector of MS LiDAR, including the coordinates and reflectance values from all channels (geometry + green + NIR case). According to the results obtained, in both levels, PTv3 demonstrates the best overall performance, achieving the highest mIoU (78.4% and 58.9%, respectively). The second-best performing model is SpUnet in LULC-L1 (mIoU = 76.3%) and SPT in LULC-L2 (mIoU = 53.6%). The confusion matrices for LULC-L1 and LULC-L2 of the best-performing model (PTv3), based on the initial MS LiDAR feature vector (geometry + green + NIR), are presented in Fig. 11, while the confusion matrices for all the different initial feature vector scenarios are provided in the Appendix (see Section 5.2). The lowest accuracy is observed for KPConv in LULC-L1 (mIoU = 70.3%) and PTv1 in LULC-L2 (mIoU = 46.7%).

At LULC-L1, PTv3 achieves the highest accuracy for ground, water, low vegetation, and “other” classes. KPConv demonstrates the best performance across the remaining classes, including medium vegetation, high vegetation, and buildings, except for bridge class, where KPConvX leads and also secures the highest mAcc.

At LULC-L2, PTv3 shows superiority, leading to the highest performance in eight classes, including road, low vegetation, façade, chimney, vehicle, electric tower, cable, and road marking. KPConvX achieves higher accuracy in segmenting medium vegetation, high vegetation, solar panels, and “other” classes. It also attains the second-best performance for road marking segmentation. Road marking and solar panel are the most challenging classes to detect using LiDAR data among the 20 classes considered. Indeed, detecting these classes solely based on geometric features is not feasible, and incorporating spectral information is required. According to Table 7, among the DL models, KPConvX achieved the highest IoU for solar panel (38.5%) and the second-highest IoU for road marking (14.3%, slightly lower than PTv3). Therefore, KPConvX provides the best overall performance for detecting these challenging classes.

On the other hand, the best segmentation accuracy for soil, water, bridges, and sport areas is achieved by SPT, demonstrating its superiority in detecting very large objects compared to other models, likely due to its superpoint-based segmentation strategy. Moreover, SPT achieves the highest OA at L2. HPF shows the highest IoU for asphalt and roofs, while the highest accuracy for segmenting poles and fences/walls is obtained by SpUnet. For vegetation segmentation in the defined three categories, KPConv is the best-performing model at L1, whereas at L2, KPConvX achieves the best performance.

HPF, a modified PTv1 featuring a double-branch architecture for LiDAR and hyperspectral data fusion, substantially outperforms the PTv1 baseline on LULC-2 by a significant margin. For LULC-1, it also improves upon PTv1, although the gain is comparatively small. This indicates that LULC-2, with its fine-grained class definitions, benefits more from specialized architectural design. Processing the two modalities separately appears advantageous for discriminating between spectrally and structurally similar classes. Compared with other models, HPF achieved moderate performance overall. One possible explanation is that HPF is specifically designed to exploit the full hyperspectral spectrum, maximizing information extraction from the input data. Additionally, the HPF design is backbone-agnostic and could be adopted to newer architectures, such as PTv3, which may further enhance performance. However, this increased architectural sophistication comes at the cost of higher computational overhead.

**Table 6** Comparison of 3D DL models for semantic segmentation of MS LiDAR data (geometry + green + NIR) at LULC-L1. The best-performing results per class are presented in bold, while the second-best results are underlined.

| Class | IoU (%) | | | | | | |
|---|---|---|---|---|---|---|---|
| | KPConv | KPConvX | SPT | HPF | PTv1 | PTv3 | SpUnet |
| Ground | 94.9 | 93.7 | <u>98.6</u> | <u>98.6</u> | 98.4 | **98.8** | **98.8** |
| Water | 22.3 | 28.8 | <u>50.5</u> | 40.4 | 37.2 | **51.1** | 47.6 |
| Low vegetation | <u>72.4</u> | 66.0 | 58.7 | 59.8 | 67.6 | **72.6** | 70.0 |
| Medium vegetation | **89.3** | <u>84.6</u> | 76.8 | 66.8 | 76.5 | 83.5 | 81.1 |
| High vegetation | **98.0** | 96.6 | 95.6 | 90.8 | 95.7 | <u>97.3</u> | 96.1 |
| Building | **91.2** | 86.5 | <u>89.9</u> | 89.4 | 85.2 | 87.2 | 88.9 |
| Bridge | 53.2 | **92.8** | 89.8 | 81.4 | 64.4 | <u>91.0</u> | 84.1 |
| Other | 41.3 | 34.7 | 38.1 | 41.0 | <u>43.4</u> | **45.8** | 43.3 |
| mIoU (%) | 70.3 | 73.0 | 74.8 | 71.1 | 71.0 | **78.4** | <u>76.3</u> |
| mAcc (%) | 83.6 | **87.8** | 84.2 | 81.2 | 78.5 | <u>85.6</u> | 84.4 |
| OA (%) | 95.2 | 94.1 | 97.8 | 97.4 | 97.6 | **98.2** | <u>98.1</u> |

**Table 7** Comparison of 3D DL models for semantic segmentation of MS LiDAR data (geometry + green + NIR) at LULC-L2. The best-performing results per class are presented in bold, while the second-best results are underlined.

| Class | KPConv | KPConvX | SPT | IoU (%)<br>HPF | PTv1 | PTv3 | SpUnet |
|---|---|---|---|---|---|---|---|
| Asphalt | 33.5 | 30.9 | 20.7 | **48.7** | 22.9 | 44.6 | 21.8 |
| Soil | 64.0 | 49.3 | **90.4** | 88.5 | 77.6 | 89.4 | 68.6 |
| Road | 33.4 | 37.4 | 35.1 | 40.1 | 13.9 | **46.9** | 18.1 |
| Water | 8.7 | 15.2 | **49.2** | 41.6 | 37.3 | 43.8 | 10.9 |
| Low vegetation | 61.6 | 68.7 | 51.7 | 63.8 | 62.8 | **72.7** | 65.9 |
| Medium vegetation | 87.6 | **87.7** | 75.8 | 82.5 | 70.5 | 84.6 | 77.1 |
| High vegetation | 97.4 | **97.6** | 95.7 | 95.6 | 94.2 | 97.0 | 95.5 |
| Roof | 90.3 | 86.7 | 90.5 | **91.3** | 83.6 | 85.9 | 85.9 |
| Façade | 42.2 | 41.6 | 37.7 | 38.4 | 38.5 | **54.7** | 53.2 |
| Chimney/roof objects | 68.5 | 55.9 | 53.0 | 63.9 | 62.0 | **70.6** | 67.7 |
| Solar panel | 35.2 | **38.5** | 25.5 | 21.1 | 15.8 | 20.2 | 11.5 |
| Vehicle | 64.4 | 60.4 | 49.8 | 43.7 | 33.7 | **66.0** | 55.8 |
| Electric tower | 80.9 | 94.0 | 92.2 | 47.1 | 96.4 | **97.6** | 86.6 |
| Cable | 91.8 | 92.4 | 93.2 | 90.1 | 93.7 | **97.1** | 95.4 |
| Pole | 53.9 | 52.7 | 47.1 | 41.4 | 39.5 | 57.1 | **61.7** |
| Bridge | 71.6 | 71.9 | **85.1** | 74.9 | 58.7 | 75.9 | 76.8 |
| Fence/wall | 29.8 | 27.4 | 31.0 | 27.9 | 29.9 | 31.1 | **35.8** |
| Sport area | 7.7 | 2.0 | **36.9** | 5.0 | 0.2 | 13.2 | 1.2 |
| Road marking | 13.5 | 14.3 | 0.6 | 9.6 | 0.3 | **14.8** | 0.2 |
| Other | 20.9 | **22.9** | 10.0 | 8.7 | 2.0 | 14.3 | 7.5 |
| mIoU (%) | 52.8 | 52.4 | 53.6 | 51.2 | 46.7 | **58.9** | 49.9 |
| mAcc (%) | 71.0 | 71.7 | 64.5 | 69.5 | 58.5 | **73.2** | 65.4 |
| OA (%) | 74.9 | 67.9 | **89.1** | 87.8 | 77.8 | 89.0 | 70.9 |

More accurate results, especially at the fine-grained LULC-L2 level, can be achieved by decreasing flight altitude. Specifically, in our dataset, road markings are very narrow, unlike those captured by Mobile/Terrestrial Laser Scanners (MLS/TLS). That said, there is currently no commercial MS MLS or TLS system available on the market. On the other hand, MLS and TLS systems are not suitable for national coverage.

From the time-efficiency perspective, since SPT segments superpoints rather than operating directly at the point level, it remains significantly more efficient than the other DL models. Indeed, training SPT takes 4.2 and 5.4 hours on a single NVIDIA RTX A6000 GPU, respectively. On the other hand, the best-performing PTv3 model requires approximately 8 hours of training on eight NVIDIA A100 SXM4 80GB GPUs. It is noteworthy that we do not report the full elapsed training time for all models in this paper, as some models are trained using multiple or different GPUs due to constraints imposed by the tuned hyperparameters and the computational limitation of NVIDIA RTX A6000.

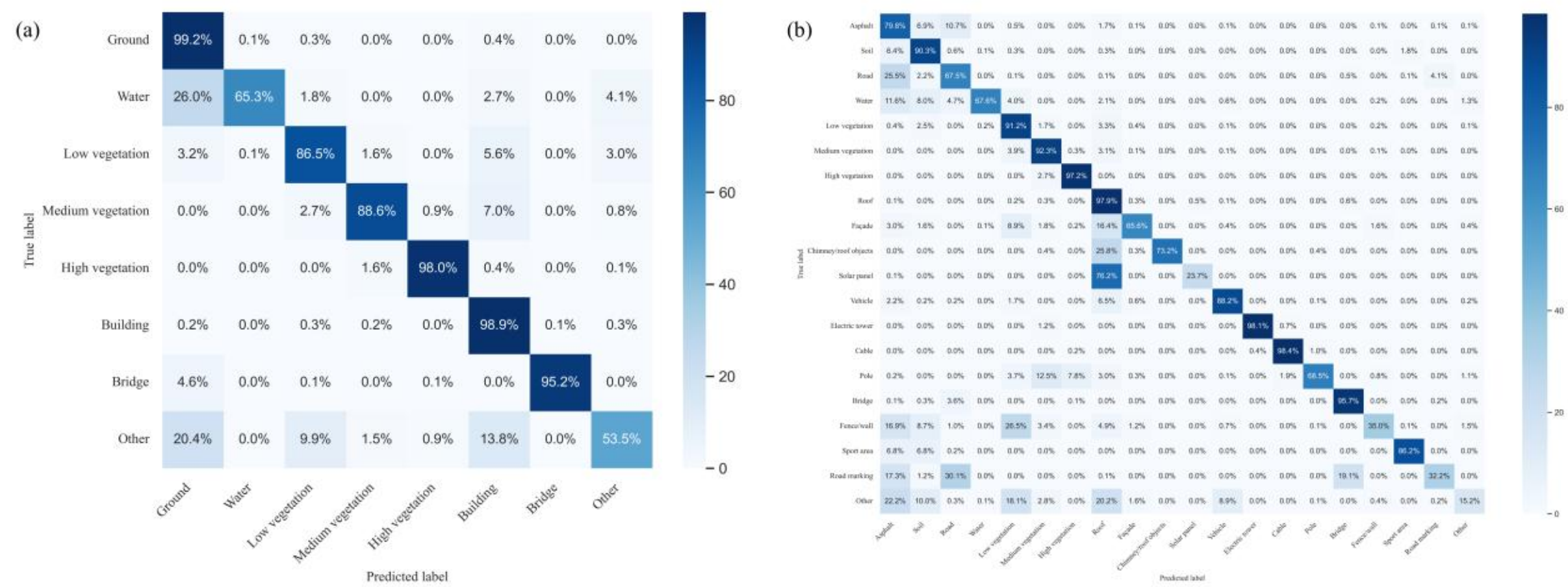

**Fig. 11.** Confusion matrices of LULC classification using MS LiDAR data at LULC-L1 (a) and LULC-L2 (b) based on the best-performing DL model (PTv3).

## 5.2. Spectral ablation study

To examine the effect of incorporating laser spectral information on LULC classification, rigorous spectral ablation studies are conducted at both the current practices level (L1) and the prospective fine-grained level (L2). These ablation studies encompass six initial feature vector scenarios: using solely geometric features; two monospectral cases (+ Green and + NIR); and three MS configurations (+ Green + NIR, + pNDVI, and + Green + NIR + pNDVI). Indeed, the effect of incorporating pNDVI is examined in two ways: using pNDVI alone and in combination with reflectance values. The PTv3, the best-performing DL model at both levels of detail (see Table 6 and Table 7), is used to conduct spectral ablation studies. Pre-trained models for geometry and the best spectral cases at both levels of detail will be available upon acceptance of the manuscript for development of future studies.

### 5.2.1. LULC-L1

The effect of the initial feature vectors on LULC-L1 classification is presented in Table 8 and Fig. 12. The conducted spectral ablation study reveals that MS LiDAR data positively influences LULC-L1 performance, increasing mIoU by 1.1 pp (+ pNDVI case), mAcc by 1.2 pp (+ Green + NIR case), and OA by 0.2 pp (+ pNDVI case) compared to using only geometric features. More specifically, MS information improves the IoU of low vegetation by 1.9 pp, medium vegetation by 4.5 pp, high vegetation by 2.8 pp, building by 1.9 pp, bridge by 4.6 pp, and "other" by 4.1 pp. Nevertheless, incorporating spectral information provides no benefit for the segmentation of ground and water classes.

On the other hand, incorporating even monospectral information is also beneficial for LULC classification, enhancing mAcc by 1.7 pp (using the NIR channel) and OA by 0.1 pp (using the green channel) compared to using only geometric features. In particular, the NIR channel boosts the IoU of medium vegetation, high vegetation, and building by 7.1 pp, 3.4 pp, and 0.6 pp, respectively, while the green channel increases the IoU of the "other" class by 1.1 pp. Although MS information is generally more beneficial for LULC-L1 than monospectral data, the results show that, for the segmentation of medium and high vegetation, the monospectral NIR channel is more effective than MS information by 2.6 pp and 0.6 pp, respectively.

A closer examination of the visual results in Fig. 12 further reveals the marginal improvement in LULC-L1 classification achieved by incorporating LiDAR spectral information. The relatively modest accuracy improvements obtained by leveraging spectral information can be attributed to the composition of LULC-L1, which includes eight classes, half of which (ground, building, bridge, and other) exhibit heterogeneous spectral characteristics due to varying surface materials. Specifically, the ground class includes asphalt, gravel, soil, roads, sports areas, and road markings, resulting in heterogeneous spectral information in both channels. Similarly, the building class includes roofs, solar panels, chimneys, and façades with diverse spectral properties. Moreover, the presence of different roof materials (e.g., clay, concrete, and metal) and varying colors and age within the same material type further increases spectral heterogeneity. For bridges, some are road bridges made of reinforced concrete, while others are wooden or asphalt structures. The "other" class, as the name suggests, includes remaining objects with highly heterogeneous spectral characteristics.

**Table 8** Spectral ablation study on LULC-L1 classification using the best-performing model (PTv3, see Table 6). The best results are highlighted in bold, and the second-best results are underlined.

| Class | IoU (%) | | | | | | Spectral benefit (pp) | |
|---|---|---|---|---|---|---|---|---|
| | Geometry | + Green | + NIR | + Green + NIR | + pNDVI | + Green + NIR + pNDVI | Mono | Multi |
| Ground | **99.2** | <u>99.1</u> | 98.9 | 98.8 | **99.2** | 99.0 | - | - |
| Water | **57.2** | 49.6 | 51.5 | 51.1 | <u>54.8</u> | 49.5 | - | - |
| Low vegetation | 73.3 | 72.9 | 71.6 | 72.6 | **75.2** | <u>75.0</u> | - | 1.9 |
| Medium vegetation | 79.0 | 83.0 | **86.1** | <u>83.5</u> | 82.7 | 80.4 | 7.1 | 4.5 |
| High vegetation | 94.5 | <u>97.6</u> | **97.9** | 97.3 | 95.2 | 94.9 | 3.4 | 2.8 |
| Building | 91.0 | 90.2 | 91.6 | 87.2 | **92.9** | <u>92.0</u> | 0.6 | 1.9 |
| Bridge | 86.4 | 86.3 | 68.7 | **91.0** | <u>87.9</u> | 71.9 | - | 4.6 |
| Other | 45.8 | 46.9 | 44.2 | 45.8 | <u>47.1</u> | **49.9** | 1.1 | 4.1 |
| mIoU (%) | 78.3 | 78.2 | 76.3 | <u>78.4</u> | **79.4** | 76.6 | - | 1.1 |
| mAcc (%) | 84.4 | 85.3 | **86.1** | <u>85.6</u> | 85.4 | 84.7 | 1.7 | 1.2 |
| OA (%) | 98.3 | <u>98.4</u> | 98.3 | 98.2 | **98.5** | 98.3 | 0.1 | 0.2 |

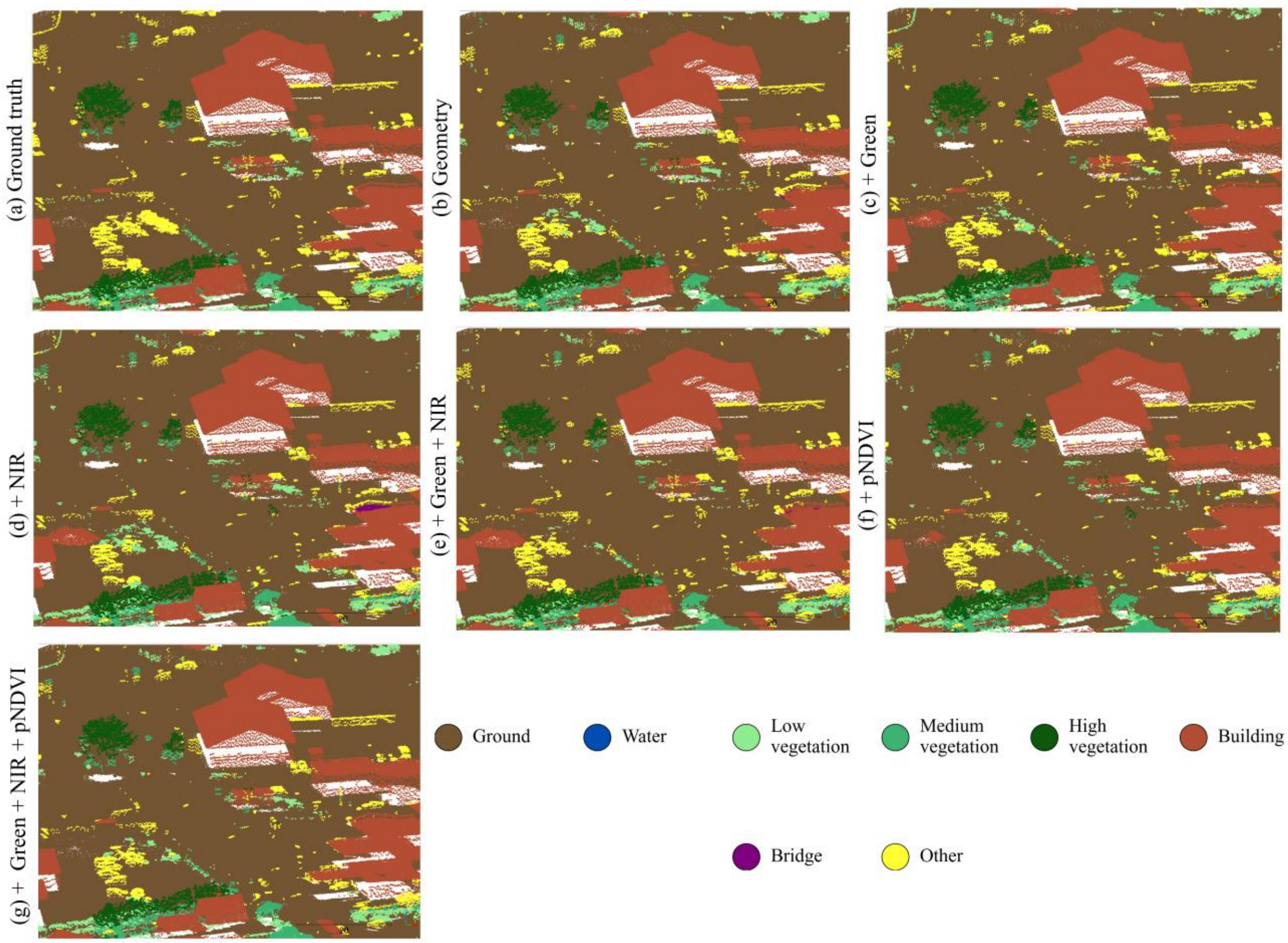

**Fig. 12.** A close-up examination of the effect of incorporating LiDAR spectral information on LULC-L1. (a) Ground truth, (b) geometry, (c) + Green, (d) + NIR, (e) + Green + NIR, (f) + pNDVI, (g) + Green + NIR + pNDVI.

In contrast, the water class in the MS point cloud is affected by significant spectral mixing. This issue arises because, during the merging of individual spectral channels to generate the MS point cloud, a global search radius is applied. A small search radius can lead to the loss of representative points due to the absence of corresponding neighbors in other channels. Conversely, a large search radius ensures the inclusion of points from larger objects (e.g., façades) but may introduce mixed and unreliable spectral information, particularly for sparsely sampled objects such as water.

Among the spectral scenarios, the +pNDVI case outperforms the others. This result demonstrates that incorporating spectral information in the form of a vegetation index can be more effective than directly using raw spectral channels (i.e., the +Green +NIR case). Furthermore, this approach offers a practical advantage by accelerating the processing of MS point clouds, as it reduces the dimensionality of the feature vector. That said, the highest IoU for bridge segmentation is achieved with the +Green +NIR case. Moreover, incorporating Green, NIR, and pNDVI yields the highest IoU for segmenting the "other" class. The confusion matrices of all the spatial and spatial–spectral initial feature scenarios at LULC-L1 are presented in Fig. B.1.

### 5.2.2. LULC-L2

The results of the spectral ablation study at LULC-L2 are reported in Table 9 and presented in Fig. 13. Like LULC-L1, both monospectral and MS information provide advantages in LULC-L2 classification. However, this advantage is more pronounced in LULC-L2 than in LULC-L1. In fact, compared to geometry case, monospectral information improves mIoU, mAcc, and OA by 5.6 pp, 1.9 pp, and 8.4 pp, respectively. The dual-wavelength MS information improves these metrics even more, by 7.8 pp, 6.0 pp, and 10.4 pp for mIoU, mAcc, and OA, respectively. The incorporation of the green and NIR channels (+Green + NIR) represents the most advantageous spectral configuration.

**Table 9** Spectral ablation study on LULC-L2 classification using the best-performing model (PTv3, see Table 7).

| Class | IoU (%) | | | | | | Spectral benefit (pp) | |
|---|---|---|---|---|---|---|---|---|
| | Geometry | + Green | + NIR | + Green + NIR | + pNDVI | + Green + NIR + pNDVI | Mono | Multi |
| Asphalt | 29.1 | 36.4 | 34.4 | 44.6 | 41.8 | **45.7** | 7.3 | 16.6 |
| Soil | 76.9 | 84.8 | 87.1 | **89.4** | **89.4** | 79.1 | 10.2 | 12.5 |
| Road | 19.7 | 21.9 | 46.4 | **46.9** | 46.6 | 38.2 | 26.7 | 27.2 |
| Water | 12.4 | 16.6 | 20.0 | **43.8** | 18.5 | 41.6 | 7.6 | 31.4 |
| Low vegetation | 70.6 | 73.0 | 72.6 | 72.7 | 74.6 | **74.8** | 2.4 | 4.2 |
| Medium vegetation | 80.2 | 83.0 | 85.1 | 84.6 | **86.6** | 85.2 | 4.9 | 6.4 |
| High vegetation | 96.3 | 97.1 | 97.6 | 97.0 | **97.8** | 97.1 | 1.3 | 1.5 |
| Roof | 87.3 | 85.9 | 88.7 | 85.9 | **88.8** | 84.8 | 1.4 | 1.5 |
| Façade | 51.4 | 56.2 | **56.4** | 54.7 | 56.3 | 56.0 | 5.0 | 4.9 |
| Chimney/roof objects | 65.3 | **74.0** | **74.0** | 70.6 | 64.4 | 67.5 | 8.7 | 5.3 |
| Solar panel | 2.4 | 17.1 | 18.6 | **20.2** | 2.7 | 2.9 | 16.2 | 17.8 |
| Vehicle | 55.9 | **66.2** | 66.1 | 66.0 | 64.5 | 58.7 | 10.3 | 10.1 |
| Electric tower | 94.4 | 94.9 | 97.0 | **97.6** | 97.1 | 97.2 | 2.6 | 3.2 |
| Cable | 96.1 | 96.8 | 96.5 | **97.1** | 96.4 | 96.4 | 0.7 | 1.0 |
| Pole | 53.3 | 53.5 | **58.3** | 57.1 | 57.2 | 57.4 | 5.0 | 4.1 |
| Bridge | **78.2** | 74.7 | 76.1 | 75.9 | 74.8 | 76.3 | - | - |
| Fence/wall | 29.1 | 29.1 | 30.1 | 31.1 | 31.4 | **32.1** | 1.0 | 3.0 |
| Sport area | 9.0 | **21.2** | 4.2 | 13.2 | 16.8 | 2.5 | 12.2 | 7.8 |
| Road marking | 1.7 | 9.4 | 13.6 | **14.8** | 11.9 | 14.1 | 11.9 | 13.1 |
| Other | 13.0 | 14.1 | 11.6 | 14.3 | 15.0 | **22.5** | 1.1 | 9.5 |
| mIoU (%) | 51.1 | 55.3 | 56.7 | **58.9** | 56.6 | 56.5 | 5.6 | 7.8 |
| mAcc (%) | 67.2 | 68.9 | 69.1 | **73.2** | 71.4 | 72.3 | 1.9 | 6.0 |
| OA (%) | 78.6 | 85.1 | 87.0 | **89.0** | 88.9 | 81.2 | 8.4 | 10.4 |

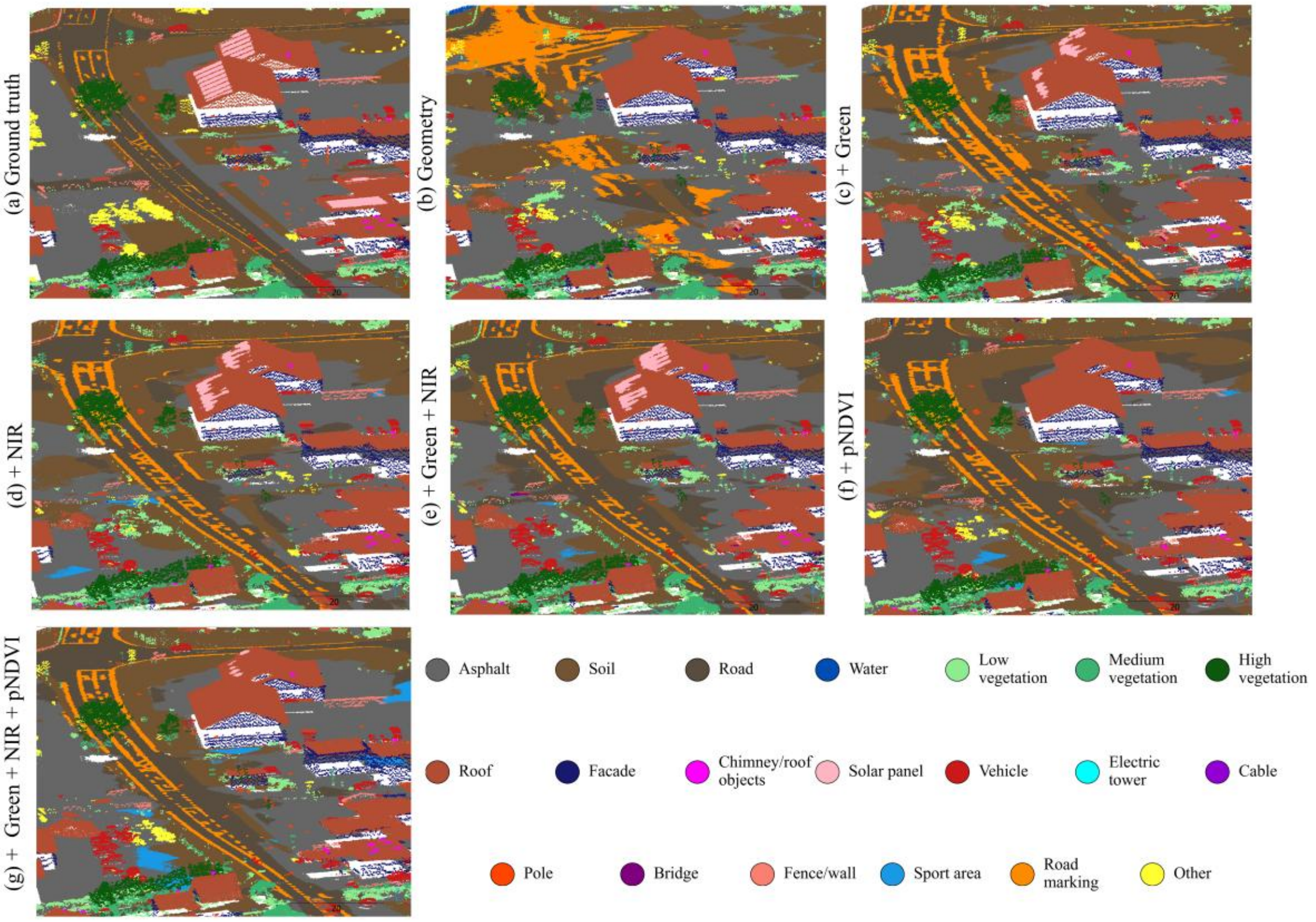

**Fig. 13.** Close-up analysis of the impact of LiDAR spectral information on LULC-L2. (a) Ground truth, (b) geometry, (c) + Green, (d) + NIR, (e) + Green + NIR, (f) + pNDVI, (g) + Green + NIR + pNDVI.

Road marking, solar panels, and sports areas are among the most challenging classes to segment using only geometric information, achieving IoU values of 1.7%, 2.4%, and 9.0%, respectively. Incorporating spectral information significantly improves their segmentation performance, increasing IoU by 13.1 pp, 17.8 pp, and 12.2 pp, respectively.
At LULC-L2, MS LiDAR provides the two most substantial improvements for the water and road classes, with IoU increases of 31.4 pp and 27.2 pp, respectively. On the other hand, incorporating monospectral information yields the greatest enhancement for road and solar panel, with IoU gains of 26.7 pp and 16.2 pp, respectively.
Notably, the spectral ablation study on the 20 classes substantiates the strong potential of MS LiDAR data for surface material detection, particularly by improving the IoU of asphalt by 16.6 pp, soil by 12.5 pp, road by 27.2 pp, and solar panels by 17.8 pp.
Among the classes at L2, spectral information provides the least benefit for segmenting cables, with improvements of 0.7 pp in the monospectral setting and 1.0 percentage points in the MS setting. Although the spectral signature of cables is relatively homogeneous, it exhibits inter-class similarity with electric towers and poles (see Fig. 8), which reduces the effectiveness of spectral information in improving segmentation performance. In addition, fences/walls, roofs, and high vegetation show less than 2 pp improvement in both the monospectral and MS cases. The limited improvement for fences/walls and roofs is related to high spectral heterogeneity caused by variations in material, color, and age. Additionally, monospectral and MS information result in improvements of 7.3 pp and 16.6 pp, respectively. That said, as mentioned earlier, the asphalt class includes some small areas of gravel. Although asphalt and gravel can appear somewhat similar in the green channel in some cases, they are usually quite distinct in the NIR channel. Such merging of underrepresented classes further results in spectral heterogeneity. On the other hand, the low enhancement for high vegetation is due to inter-class spectral similarity with low and medium vegetation as shown in Fig. 8.
Road marking is identified as the most challenging class to detect. More specifically, across the six initial feature configurations, the highest IoU for road marking (14.8%) is achieved by incorporating both the green and NIR channels (+ Green + NIR). The visual results in Fig. 13 demonstrate that, in contrast to the geometry-only case, which leads to noticeable mis-segmentation of road markings as either road or asphalt, the inclusion of monospectral or MS LiDAR information significantly improves road marking detection. In fact, most road marking locations are correctly identified in spatial-spectral cases. However, the segmented road marking regions are considerably wider than their actual narrow extents, which contributes to the relatively low achieved accuracy for this class. Additionally, some road markings located on road bridges are mis-classified as bridge surfaces, leading to further segmentation errors.
While incorporating monospectral or MS information is beneficial for all classes in LULC-L2, bridge is the only class for which geometric information alone appears sufficient based on the obtained results. Besides, our spectral ablation study reveals that certain wavelengths are not suitable for detecting some classes. For instance, the NIR wavelength negatively affects the detection of sport area, while the green wavelength has a detrimental impact on detecting the roof class. Moreover, incorporating MS information yields smaller improvements compared to monospectral data for some classes, including façade, chimney/roof objects, vehicle, pole, and sport area. Confusion matrices at LULC-L2 for different spatial and spatial–spectral initial features are shown in Fig. B.2.
Our experiments substantiate that spectral features should be carefully selected with respect to the target objects for detection. In fact, richer spectral information does not necessarily yield the optimal feature representation. For example, as shown in Fig. 13, the inclusion of pNDVI reduces the detection performance of solar panels, bringing it down to a level comparable to the geometry-only case (see also Table 9).

# 6. Discussion

## 6.1. Variation in spectral information

Unlike photogrammetric point cloud classification, laser reflectance values are highly sensitive features in MS LiDAR point clouds. Distinctive materials reflectance values play a crucial role in MS LiDAR-based LULC classification. In fact, even the same fine-grained classes can exhibit variations in color, age, and pattern, resulting in high inter-class spectral variability and making spatial–spectral classification challenging. For instance, as illustrated in Fig. 14, similar roof materials in our case study have different colors and patterns, with some roofs being new and others old. These factors alter the range of reflectance values of similar roof materials. Another example is fences. During our field survey, we observed that, in addition to geometrical differences, fences can be made of different materials and colors, with varying ages (see Fig. 15). A similar situation occurs with vehicles, which may appear in different colors and materials.
Moreover, reflectivity is very sensitive to moisture. Although variability in spectral values on certain land surfaces due to moisture differences can be beneficial for applications such as soil moisture estimation, it can pose challenges for LULC classification and object detection. In our case study, we observe that the tennis court exhibits a broadly similar spectral histogram to that of the football field due to the effects of surface moisture, senescent surface conditions, and surface roughness or compaction on reflectance values (see Fig. 16).Thus, although RGB orthophotos and photogrammetric point clouds easily distinguish between tennis courts and football field, in our case the MS LiDAR data does not provide the same level of discrimination in separating football fields from tennis courts.
Furthermore, some classes in LULC classification are defined based on the human purpose or function of the land, rather than solely on physical cover. For example, grass and football fields share the same physical surface cover and, consequently, exhibit similar spectral information. This inter-class similarity in laser spectral information while the geometric features are also similar, poses a challenge for their correct segmentation using MS LiDAR data. On the other hand, seasonal conditions may alter laser reflectance, as demonstrated by Karila et al. (2018). Their experiments utilized three MS LiDAR datasets acquired with the Optech Titan system at different times (May, June, and August). The results showed that the dataset collected in leaf-off conditions (May) produced more complete road representations compared to those collected during leaf-on conditions. Furthermore, they concluded that all datasets were suitable for LULC classification (including buildings, trees, asphalt, gravel, rocky areas, and low vegetation). However, there were minor differences in classification accuracy between the resulting LULC maps, with the most accurate (by 2.6 pp in OA) achieved using the June dataset.

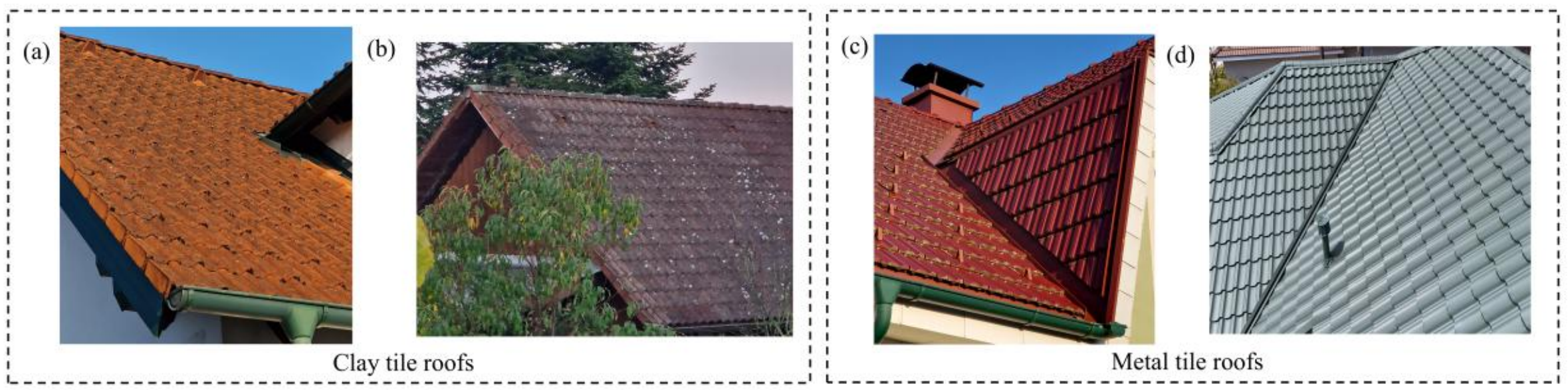

**Fig. 14.** Different colors, ages, and patterns of the roofs in our case study collected during the field survey. Panels (a) and (b) denote clay tile roofs, while panels (c) and (d) denote metal roofs.

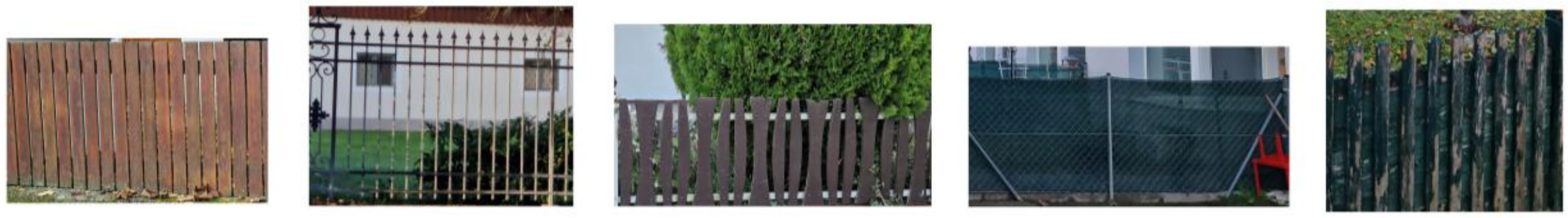

**Fig. 15.** Great variability of shapes, materials, and colors of fences in the study areas.

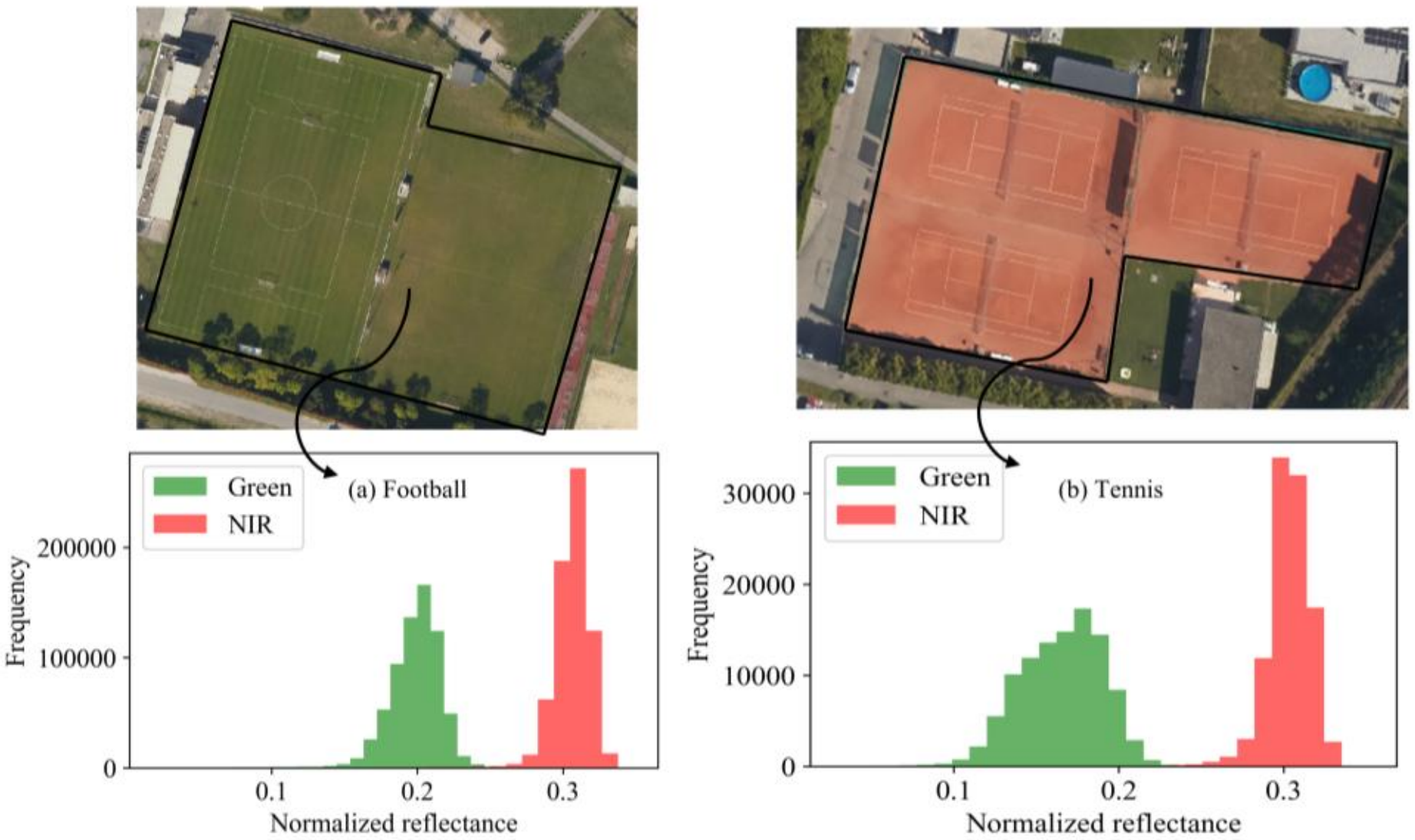

**Fig. 16.** Spectral histograms for the football field (a) and tennis courts (b) in the study area. Reflectance values are normalized to [0, 1].

### 6.2 Laser shadow-like effect

During the preparation of the Loosdorf-MSL data, a shadow-like effect is observed on the ground in both the green and NIR channels, as illustrated in Fig. 17. The shadowed ground points are caused by return echoes from an adjacent electric tower. While these shadowed points are associated with higher return numbers (more than one), some of them in the green channel also originate from first-return points. Furthermore, as shown in Fig. 17c, the shadow directions differ between the green and NIR channels. Although these shadowed points constitute a very small portion of the dataset and therefore have almost no effect on LULC classification, the correction of such radiometric errors should be investigated in future work to improve LULC classification with more reliable spectral information. In this study, radiometric correction is considered beyond the scope of the work. However, to fully exploit the potential of MS LiDAR data, the radiometric information of the raw data should be corrected not only for range effects, but also with respect to the incidence angle and potential laser shadow-like effects.

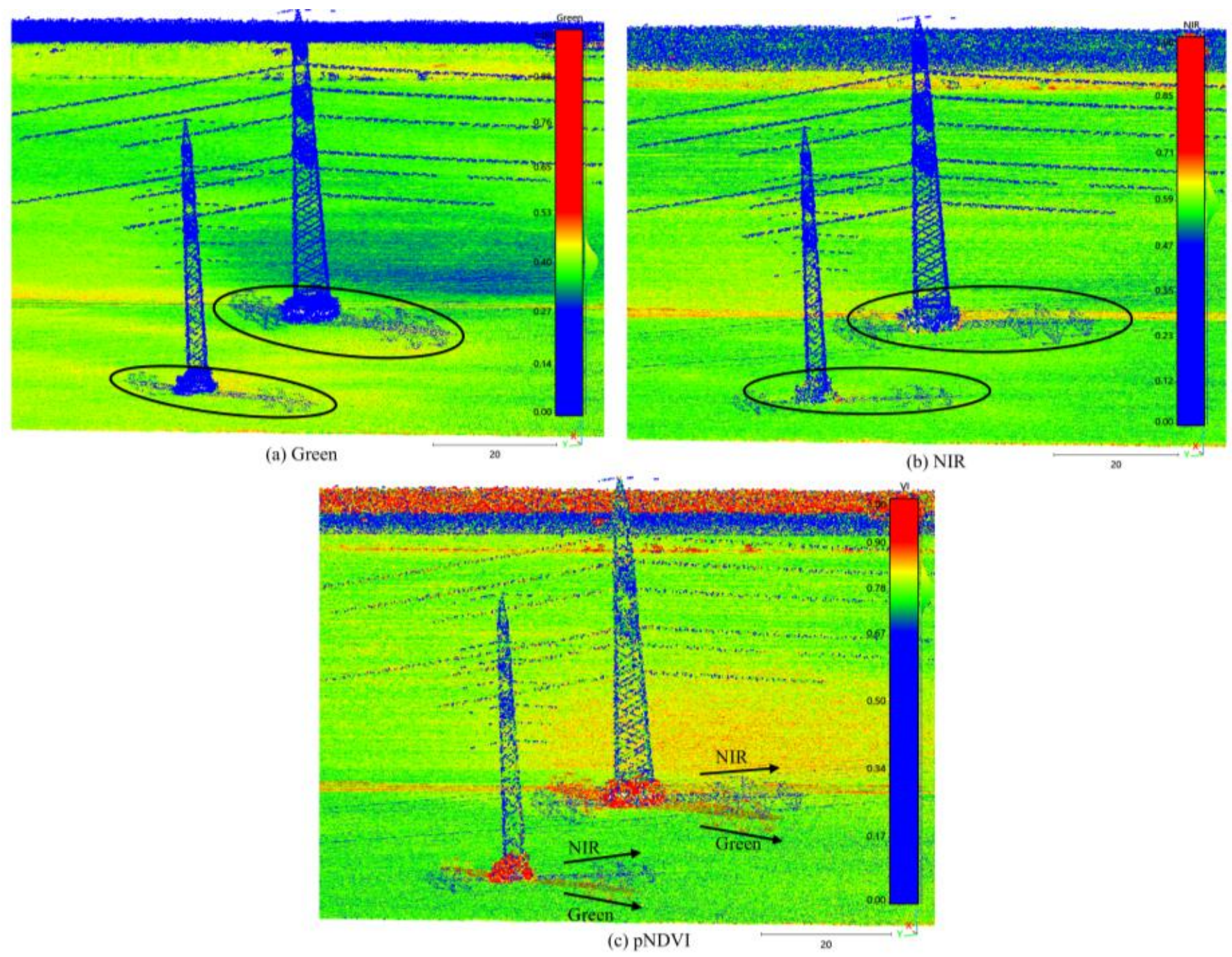

**Fig. 17.** Shadow-like effect in laser spectral information: (a) green channel, (b) NIR channel, and (c) pNDVI. Spectral values are normalized to the range [0, 1].

### 6.3. Vegetation separation

The current approach to separating vegetation, as followed by NMCAs and many studies like TerLiDAR dataset (Carós et al., 2026), relies on applying height thresholds to vegetation points without considering their spatial relationships or a full 3D understanding. As a result, a single tree may be classified simultaneously as low, medium, and high vegetation. This reduces the reliability and usefulness of the final classification product. With the current trend toward developing digital twins of urban trees and forests, there is a need to separate vegetation into low, medium, and high categories based on a full 3D understanding, rather than relying solely on height thresholds, to avoid the mentioned problem. This full 3D vegetation separation should be followed in prospective studies.

## 7. Conclusions

This study examines the potential benefits of LiDAR reflectance and advanced MS LiDAR technology for LULC classification, specifically to meet the current and prospective needs of NMCAs. In this regard, since currently there is no established standard for ALS–based LULC classification, this paper first reports the NMCAs current and prospective LULC classification schemes derived from a questionnaire conducted with the participation of European NMCAs, aimed to ensure consistency and compatibility among ALS-based national mapping products. Secondly, we present the first 3D MS LiDAR benchmark dataset for LULC classification tailored to NMCAs schemes, Loosdorf-MSL. This dataset not only facilitates future development of tailored artificial intelligent models that can handle modern and high-dimensional MS LiDAR data, but we also hope that thanks to being prepared at two levels of detail in accordance with NMCAs schemes (current practices and fine-grained prospective), it will serve as a dominant benchmark dataset in ALS-based LULC classification domain within the community, even for monochromatic LiDAR-based LULC classification. Moreover, by including 20 detailed classes, the Loosdorf-MSL dataset contributes to enhancing the diversity of point clouds semantic segmentation benchmarks and consequently advancing 3D scene understanding in remote sensing applications.

Subsequently, the advantages of MS LiDAR data for LULC classification are examined by benchmarking seven SOTA DL models, including one specifically designed for processing MS point clouds (HPF). The benchmarking of DL models substantiates the lack of appropriate DL architectures that fully exploit both spatial and spectral information of MS LiDAR data.

To examine the impact of MS LiDAR systems on both the current and future needs of NMCAs, LULC classification is performed at two levels of detail, comprising eight (LULC-L1) and 20 (LULC-L2) classes, respectively. Our experiments reveal the superior performance of PTv3 among the implemented DL models at both levels of detail, achieving mIoU values of 79.4% for LULC-L1 and 58.9% for LULC-L2. Given that processing MS LiDAR data is inherently time-consuming, our experiments further highlight the need to develop more time-efficient DL models, like SPT, in the future for efficient MS LiDAR data processing.

Furthermore, two rigorous spectral ablation studies are conducted, encompassing six different spatial and spatial–spectral initial feature vector scenarios input to the DL models. The results show that incorporating geometric features together with pNDVI provided the optimal configuration for LULC-L1, marginally improving mIoU by 1.1 pp compared to only using geometric information of LiDAR data. At LULC-L2, the best-performing initial feature vector is the direct use of the green and NIR channels, resulting in an mIoU improvement of 7.8 pp. Remarkably, our experiments demonstrate that use of reflectance and MS LiDAR becomes increasingly beneficial as the number of classes increases. In LULC-L2 classification, MS LiDAR data facilitates the discrimination of additional objects such as road markings, solar panels, and soil, owing to their distinct surface materials.

This study demonstrated the efficacy of MS LiDAR systems for LULC classification at the national scale, offering a single-source data solution for fine-grained LULC mapping that overcomes the challenges associated with data fusion. By showing a substantial improvement in LULC classification accuracy using even monospectral information, this study highlights the potential of utilizing laser radiometric information provided by conventional LiDAR systems – an area that has received limited attention to date while spectral information is already available in all LiDAR data. Most notably, involving laser spectral information facilitates material identification and infrastructure monitoring, such as identifying roof materials and distinguishing between paved and dirt roads. Furthermore, by providing insights into the classification schemes used by NMCAs, this study supports the effective adoption of MS LiDAR data for detailed LULC classification according to NMCAs' requirements.

The only currently available commercial MS LiDAR system is utilized in this study which is a dual-wavelength system; other MS LiDAR systems should be explored in future work. Furthermore, the integration of MS LiDAR data with photogrammetric point clouds (provided in our Loosdorf-MSL benchmark datasets) to generate higher-dimensional datasets can be explored in future studies.

Additionally, this study examined only one spectral index (pNDVI); other spectral indices should be investigated in future research. Spatial–spectral postprocessing remains an underexplored topic in MS LiDAR data analysis, and since it could significantly improve the accuracy of MS LiDAR-based LULC classification, it warrants further studies. In this study, the reflectance values used are only range corrected. Therefore, future research should focus on full radiometric calibration with respect to both range and incidence angle and assess its impact on LULC classification performance.

## Credit authorship contribution statement

**Narges Takhtkeshha:** Writing – original draft, Writing – review & editing, Visualization, Software, Methodology, Investigation, Data curation, Conceptualization. **Aldino Rizaldy:** Writing – review & editing, Methodology, Data curation. **Markus Hollaus:** Writing – review & editing, Supervision. **Fabio Remondino:** Writing – review & editing, Supervision, Resources, Funding acquisition. **Juha Hyyppä:** Writing – review & editing, Supervision, Resources, Funding acquisition. **Gottfried Mandlburger:** Writing – review & editing, Supervision, Resources, Project administration, Funding acquisition.

## Acknowledgement

The authors sincerely thank EuroSDR and all its national mapping agency delegates for their contributions to our questionnaire. The authors also express their gratitude to *RIEGL* for supporting data acquisition.

## Funding

This research is funded by EuroSDR and Research Council of Finland project 359203.

## Appendix A. Hybrid adjustment validation

The conducted hybrid adjustment is evaluated using Ground Control Points (GCPs) collected through GPS (Global Positioning System) field measurements. In this regard, the height deviations between the photogrammetric point clouds and each LiDAR channel are compared before and after the hybrid adjustment (Table A.1). In addition, the height deviations between the GCPs and each LiDAR channel are reported in Table A.2. Moreover, the height deviations between the GCPs and the photogrammetric point clouds are presented in Table A.3. The obtained results confirm the considerable positive impact of the hybrid adjustment on the georeferencing and co-alignment of MS LiDAR point clouds and photogrammetric point clouds.

**Table A.1** Height deviations between the photogrammetric and laser point clouds before and after hybrid adjustment.

| GCP | Photogrammetric-green laser channel (m) | | Photogrammetric-NIR laser channel (m) | |
|---|---|---|---|---|
| | Before | After | Before | After |
| P1 | -0.23 | 0.04 | -0.23 | 0.04 |
| P2 | -0.20 | 0.01 | -0.20 | 0.00 |
| P3 | -0.21 | 0.03 | -0.20 | 0.04 |
| P4 | -0.19 | 0.03 | -0.19 | 0.02 |

**Table A.2** Height deviations between GCPs and laser point clouds before and after hybrid adjustment.

| GCP | GCP-green laser channel (m) | | GCP-NIR laser channel (m) | |
|---|---|---|---|---|
| | Before | After | Before | After |
| P1 | -0.13 | -0.08 | -0.12 | -0.08 |
| P2 | -0.11 | -0.06 | -0.11 | -0.06 |
| P3 | -0.09 | -0.05 | -0.09 | -0.06 |
| P4 | -0.09 | -0.04 | -0.08 | -0.05 |

**Table A.3** Height deviations between GCPs and photogrammetric point clouds before and after hybrid adjustment

| GCP | GCP-photogrammetric point clouds (m) | |
|---|---|---|
| | Before | After |
| P1 | 0.11 | -0.04 |
| P2 | 0.09 | -0.07 |
| P3 | 0.12 | -0.02 |
| P4 | 0.10 | -0.06 |

## Appendix B. confusion matrices

Confusion matrices for different spatial and spatial–spectral initial feature scenarios are provided in Fig. B.1 and Fig. B.2 for LULC-L1 and LULC-L2, respectively.

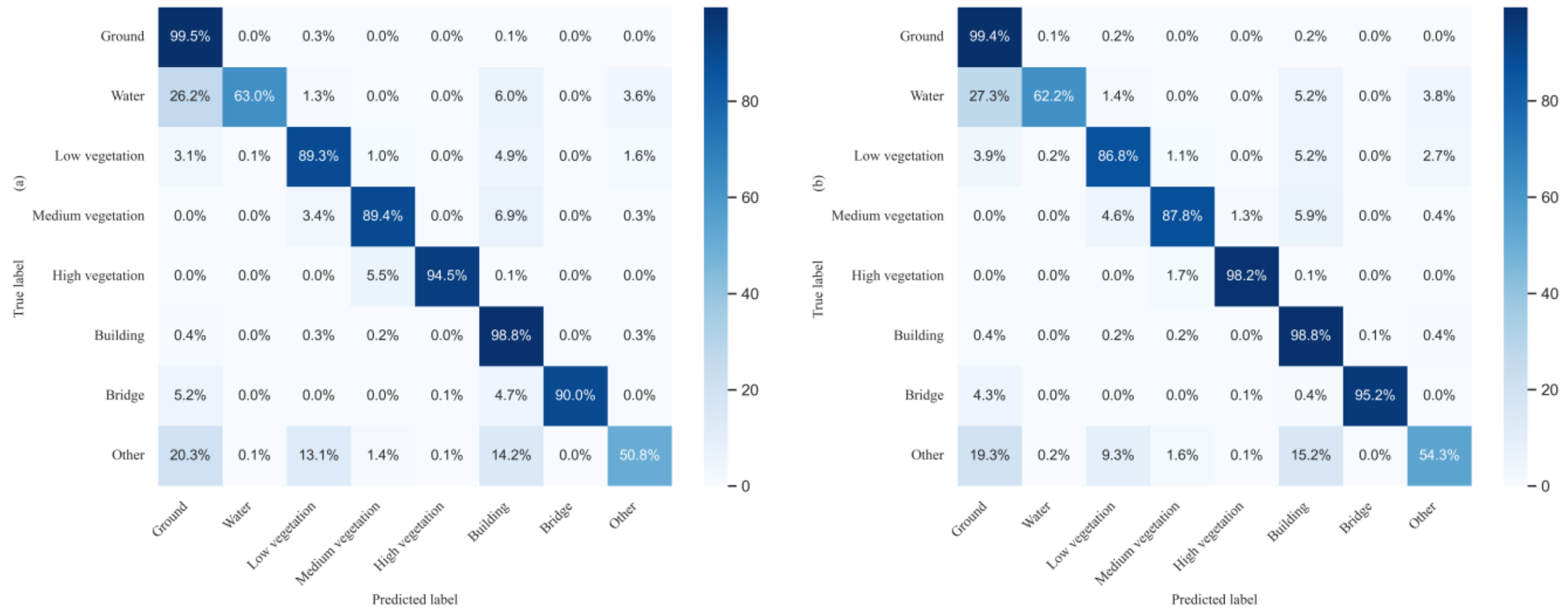

(c)

| True label \ Predicted label | Ground | Water | Low vegetation | Medium vegetation | High vegetation | Building | Bridge | Other |
|---|---|---|---|---|---|---|---|---|
| Ground | 99.2% | 0.1% | 0.4% | 0.0% | 0.0% | 0.2% | 0.1% | 0.0% |
| Water | 25.0% | 66.4% | 2.0% | 0.0% | 0.0% | 3.7% | 0.0% | 2.8% |
| Low vegetation | 2.7% | 0.1% | 88.5% | 4.0% | 0.0% | 3.2% | 0.0% | 1.4% |
| Medium vegetation | 0.0% | 0.0% | 1.2% | 94.5% | 1.6% | 2.5% | 0.0% | 0.2% |
| High vegetation | 0.0% | 0.0% | 0.0% | 1.2% | 98.7% | 0.0% | 0.0% | 0.0% |
| Building | 0.3% | 0.0% | 0.3% | 0.6% | 0.0% | 98.1% | 0.3% | 0.3% |
| Bridge | 4.6% | 0.0% | 0.1% | 0.0% | 0.1% | 0.0% | 95.2% | 0.0% |
| Other | 20.0% | 0.1% | 16.4% | 3.6% | 0.2% | 11.6% | 0.0% | 48.2% |

(d)

| True label \ Predicted label | Ground | Water | Low vegetation | Medium vegetation | High vegetation | Building | Bridge | Other |
|---|---|---|---|---|---|---|---|---|
| Ground | 99.2% | 0.1% | 0.3% | 0.0% | 0.0% | 0.4% | 0.0% | 0.0% |
| Water | 26.0% | 65.3% | 1.8% | 0.0% | 0.0% | 2.7% | 0.0% | 4.1% |
| Low vegetation | 3.2% | 0.1% | 86.5% | 1.6% | 0.0% | 5.6% | 0.0% | 3.0% |
| Medium vegetation | 0.0% | 0.0% | 2.7% | 88.6% | 0.9% | 7.0% | 0.0% | 0.8% |
| High vegetation | 0.0% | 0.0% | 0.0% | 1.6% | 98.0% | 0.4% | 0.0% | 0.1% |
| Building | 0.2% | 0.0% | 0.3% | 0.2% | 0.0% | 98.9% | 0.1% | 0.3% |
| Bridge | 4.6% | 0.0% | 0.1% | 0.0% | 0.1% | 0.0% | 95.2% | 0.0% |
| Other | 20.4% | 0.0% | 9.9% | 1.5% | 0.9% | 13.8% | 0.0% | 53.5% |

(e)

| True label \ Predicted label | Ground | Water | Low vegetation | Medium vegetation | High vegetation | Building | Bridge | Other |
|---|---|---|---|---|---|---|---|---|
| Ground | 99.6% | 0.0% | 0.2% | 0.0% | 0.0% | 0.1% | 0.0% | 0.0% |
| Water | 29.2% | 63.4% | 1.1% | 0.0% | 0.0% | 3.3% | 0.0% | 3.0% |
| Low vegetation | 4.3% | 0.2% | 89.9% | 1.3% | 0.0% | 3.3% | 0.0% | 1.0% |
| Medium vegetation | 0.0% | 0.0% | 3.3% | 93.2% | 0.0% | 3.1% | 0.0% | 0.4% |
| High vegetation | 0.0% | 0.0% | 0.0% | 4.7% | 95.3% | 0.0% | 0.0% | 0.0% |
| Building | 0.5% | 0.0% | 0.4% | 0.4% | 0.0% | 98.3% | 0.1% | 0.3% |
| Bridge | 5.5% | 0.0% | 0.1% | 0.0% | 0.1% | 1.9% | 92.4% | 0.0% |
| Other | 22.2% | 0.4% | 12.1% | 1.7% | 0.2% | 12.4% | 0.0% | 51.0% |

(f)

| True label \ Predicted label | Ground | Water | Low vegetation | Medium vegetation | High vegetation | Building | Bridge | Other |
|---|---|---|---|---|---|---|---|---|
| Ground | 99.4% | 0.0% | 0.3% | 0.0% | 0.0% | 0.1% | 0.1% | 0.0% |
| Water | 37.4% | 53.0% | 1.5% | 0.0% | 0.0% | 4.9% | 0.0% | 3.2% |
| Low vegetation | 2.8% | 0.0% | 91.2% | 1.0% | 0.0% | 3.8% | 0.0% | 1.2% |
| Medium vegetation | 0.0% | 0.0% | 3.8% | 90.2% | 0.0% | 5.7% | 0.0% | 0.3% |
| High vegetation | 0.0% | 0.0% | 0.0% | 5.0% | 95.0% | 0.1% | 0.0% | 0.0% |
| Building | 0.3% | 0.0% | 0.3% | 0.2% | 0.0% | 98.7% | 0.0% | 0.4% |
| Bridge | 4.2% | 0.0% | 0.0% | 0.0% | 0.1% | 1.5% | 94.2% | 0.0% |
| Other | 21.0% | 0.0% | 9.9% | 1.0% | 0.1% | 12.3% | 0.0% | 55.7% |

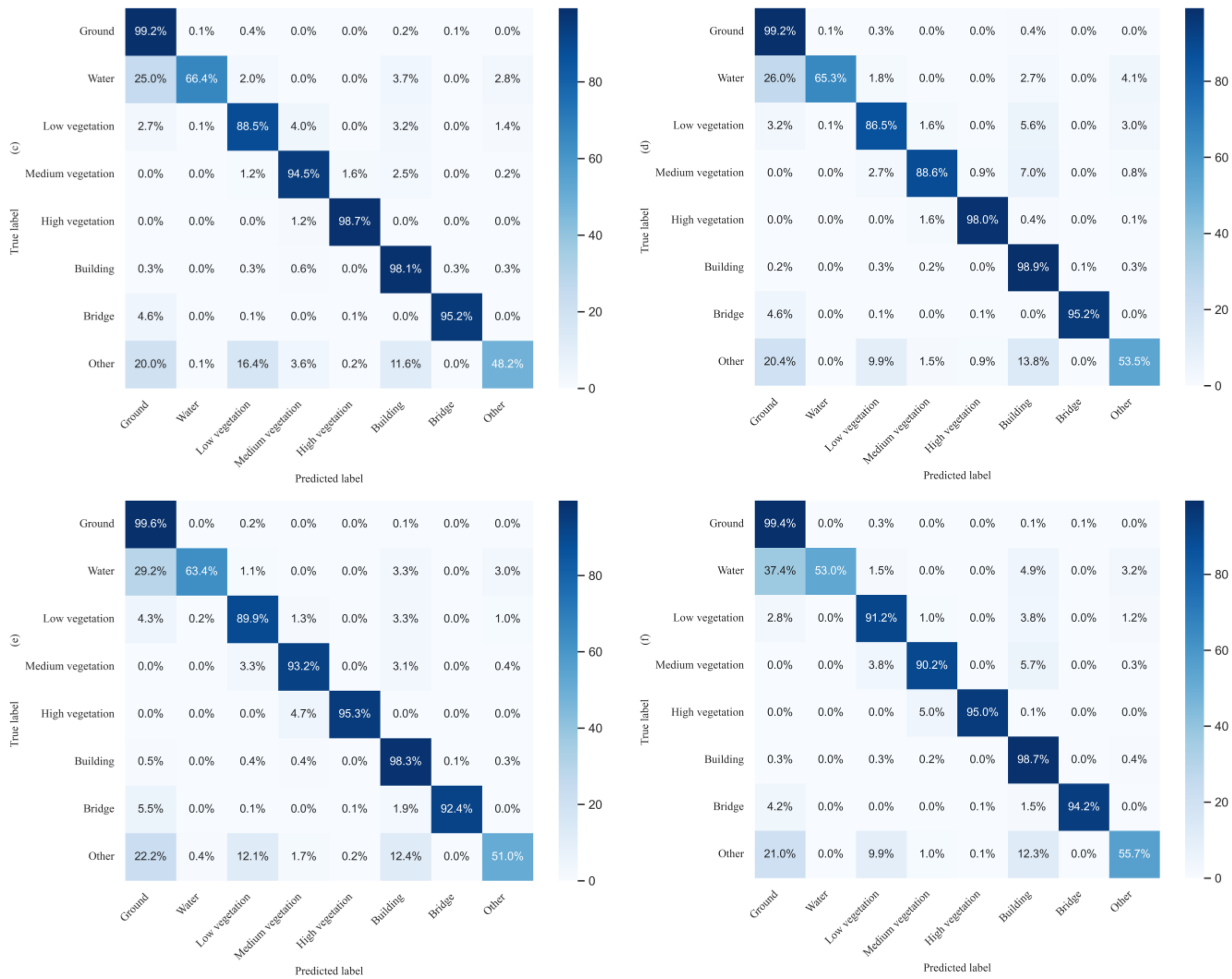

**Fig. B.1.** Confusion matrices of different spatial and spatial-spectral initial features for LULC-L1 classification using best performing DL model (PTv3); (a): geometry, (b): + Green; (c): + NIR; (d): + Green + NIR, (e) + pNDVI, (f) + Green + NIR + pNDVI.

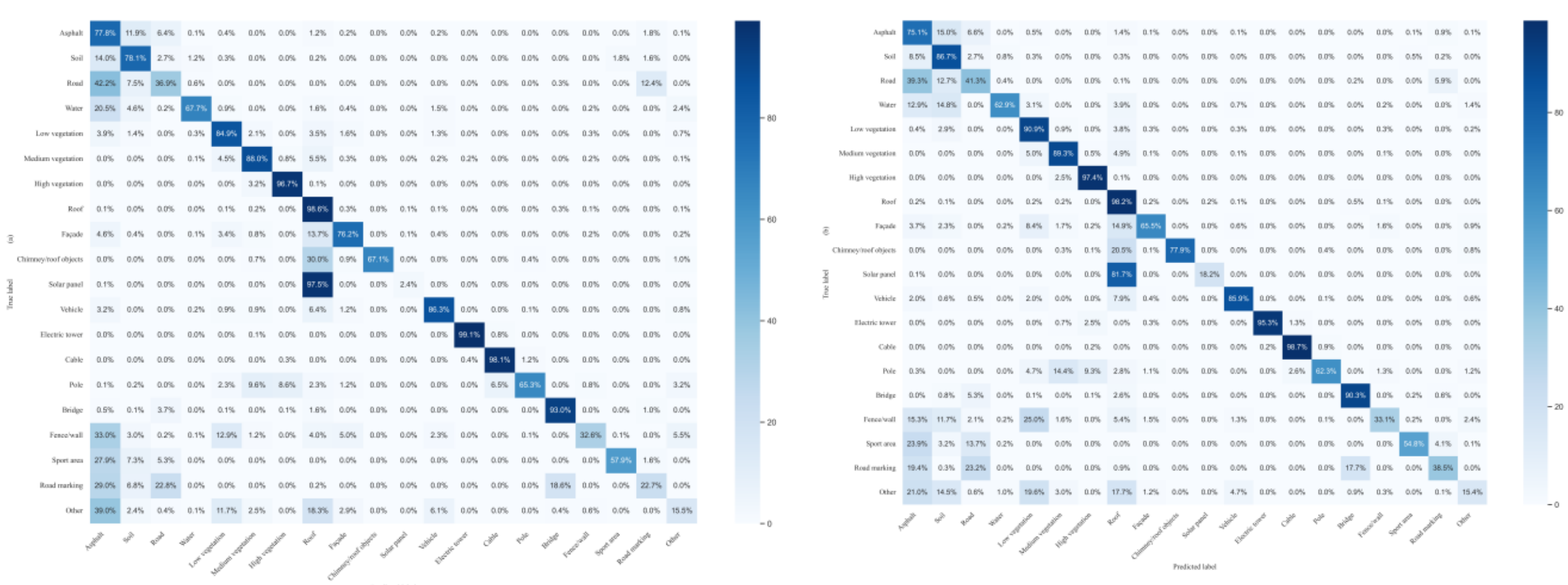

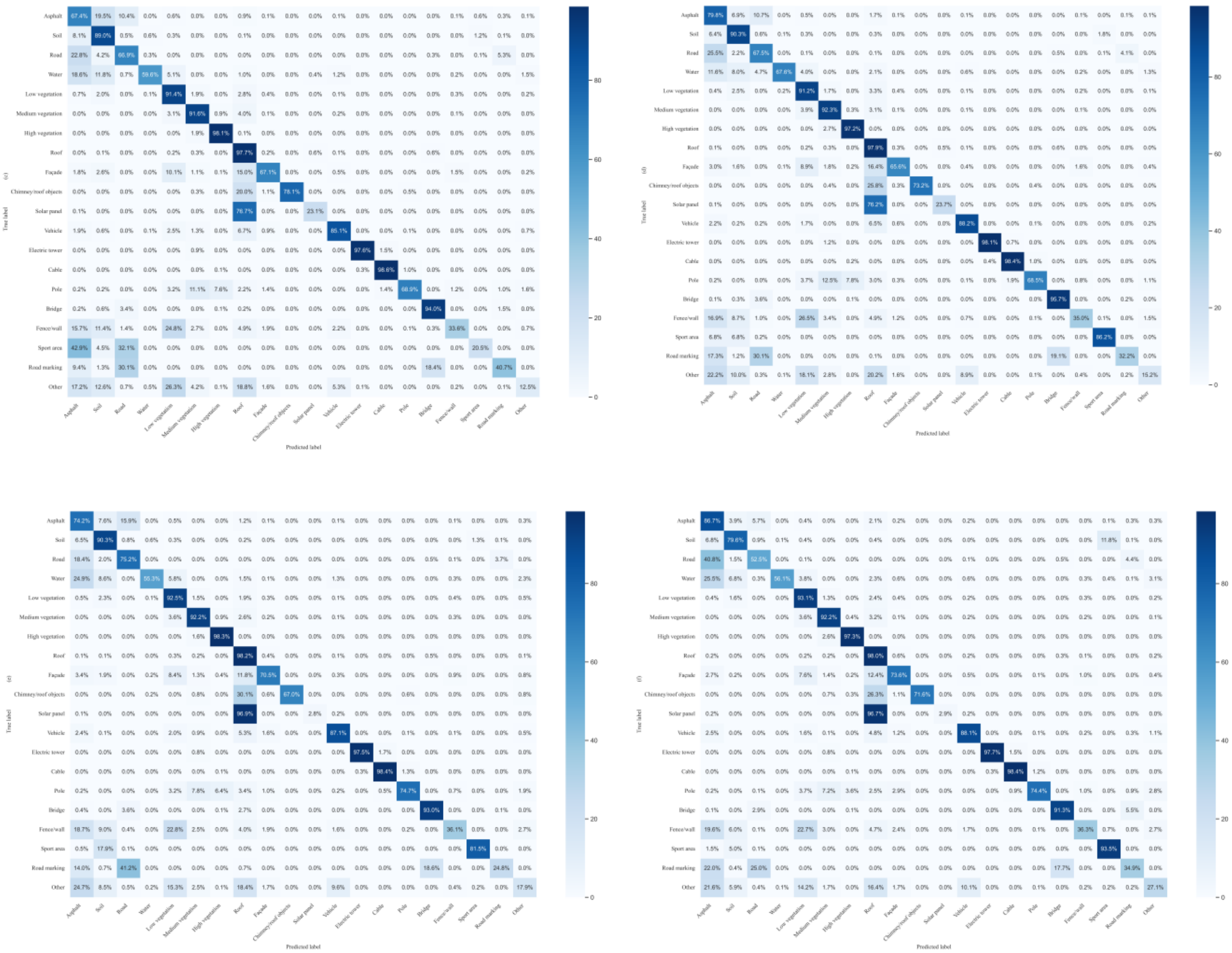

**Fig. B.2.** Confusion matrices of different spatial and spatial–spectral input features for LULC-L2 classification using the best-performing DL model (PTv3): (a) geometry, (b) + Green , (c) + NIR band, (d) + Green + NIR bands, (e) + pNDVI, and (f) + Green + NIR + pNDVI.

## References

Bayrak, O. C., Ma, Z., Farella, E., Remondino, F., & Uzar, M., 2024. ESTATE: A Large Dataset of Under-Represented Urban Objects for 3D Point Cloud Classification. The International Archives of the Photogrammetry, Remote Sensing and Spatial Information Sciences, XLVIII-2-2024. https://doi.org/10.5194/isprs-archives-XLVIII-2-2024-25-2024.

Bayrak, O. C., Remondino, F., & Uzar, M., 2023. A new dataset and methodology for urban-scale 3D point cloud classification. The International Archives of the Photogrammetry, Remote Sensing and Spatial Information Sciences, XLVIII-1/W3-2023. https://doi.org/10.5194/isprs-archives-XLVIII-1-W3-2023-1-2023.

Carós, M., Just, A., Seguí, S., & Vitrià, J., 2026. Enhancing point cloud semantic segmentation via scalable domain adaptation with LoRA-enabled PointNet++. ISPRS Open Journal of Photogrammetry and Remote Sensing, 19, 100119. https://doi.org/10.1016/j.ophoto.2026.100119.

Chen, Z., 2018. Convolutional Neural Networks for Land-cover Classification Using Multispectral Airborne Laser Scanning Data. University of Waterloo: Waterloo, ON, Canada.

Gao, L., Liu, Y., Chen, X., Liu, Y., Yan, S., Zhang, M., 2024. CUS3D: A new comprehensive urban-scale semantic-segmentation 3D benchmark dataset. Remote Sensing 16(6), 1079.

Ghirardeau-Montaut, D., 2021. CloudCompare 3D point cloud and mesh processing software open source project. https://www.danielgm.net/cc/, 2.2.21.

Glira, P., Pfeifer, N., Mandlburger, G., 2019. Hybrid orientation of airborne lidar point clouds and aerial images. In: ISPRS Annals of the Photogrammetry, Remote Sensing and Spatial Information Sciences, Geospatial Week 2019, Enschede, NL.

Graham, B., Engelcke, M., & Maaten, L., 2018. 3D semantic segmentation with submanifold sparse convolutional networks. Proc. IEEE/CVF Conf. Comput. Vis. Pattern Recognit., 9224–9232. https://doi.ieeecomputersociety.org/10.1109/CVPR.2018.00961.

Han, X., Liu, C., Dong, Z., & Yang, B., 2024. WHU Urban3D: 3D urban scale point cloud dataset and deep learning baseline for semantic & instance segmentation. ISPRS Journal of Photogrammetry and Remote Sensing.

Han, X., Liu, C., Zhou, Y., Tan, K., Dong, Z., & Yang, B., 2024. WHU-Urban3D: an urban scene LiDAR point cloud dataset for semantic instance segmentation. *ISPRS Journal of Photogrammetry and Remote Sensing*, 209, 500–513.

Hu, Q., et al., 2020. RandLA-Net: Efficient Semantic Segmentation of Large-Scale Point Clouds. In 2020 IEEE/CVF Conference on Computer Vision and Pattern Recognition (CVPR), 11105–11114. https://doi.org/10.1109/CVPR42600.2020.01112.

Jing, Z., Guan, H., Zhao, P., Li, D., Yu, Y., Zang, Y., Wang, H., & Li, J., 2021. Multispectral lidar point cloud classification using SE-PointNet++. Remote Sens., 13, 2516.

Kaasalainen, S., Lindroos, T., Hyyppä, J., 2007. Toward Hyperspectral Lidar: Measurement of Spectral Backscatter Intensity With a Supercontinuum Laser Source. IEEE Geosci. Remote Sens. Lett. 4, 211–215. https://doi.org/10.1109/LGRS.2006.888848.

Karila, K., Matikainen, L., Litkey, P., Hyyppä, J., & Puttonen, E., 2018. The effect of seasonal variation on automated land cover mapping from multispectral airborne laser scanning data. International Journal of Remote Sensing, 40, 1–19. https://doi.org/10.1080/01431161.2018.1528023.

Kölle, M., Laupheimer, D., Schmohl, S., Haala, N., Rottensteiner, F., & Wegner, J., 2021. The Hessigheim 3D (H3D) benchmark on semantic segmentation of high-resolution 3D point clouds and textured meshes from UAV LiDAR and Multi-View-Stereo. ISPRS Open Journal of Photogrammetry and Remote Sensing, 1, 100001. https://doi.org/10.1016/j.ophoto.2021.100001.

Li, D., Shen, X., Guan, H., Yu, Y., Wang, H., Zhang, G., & Li, J., 2022. AGFP-Net: Attentive geometric feature pyramid network for land cover classification using airborne multispectral LiDAR data. Int. J. Appl. Earth Obs. Geoinf., 108, 102723. https://doi.org/10.1016/j.jag.2022.102723.

Liao, Y., Xie, J., & Geiger, A., 2022. KITTI 360: A Novel Dataset and Benchmarks for Urban Scene Understanding in 2D and 3D. IEEE Transactions on Pattern Analysis and Machine Intelligence. https://doi.org/10.1109/TPAMI.2022.3179507.

Mandlburger, G., Otepka, J., Wagner, W., Karel, W., & Pfeifer, N., 2009. Orientation and processing of airborne laser scanning data (OPALS) - Concept and first results of a comprehensive ALS software. https://www.researchgate.net/publication/237437116.

Mikael Reichler, Josef Taher, Petri Manninen, Harri Kaartinen, Juha Hyyppä, & Antero Kukko, 2024. Semantic segmentation of raw multispectral laser scanning data from urban environments with deep neural networks. ISPRS Open Journal of Photogrammetry and Remote Sensing, 12, 100061. https://doi.org/10.1016/j.ophoto.2024.100061.

Morsy, S., Shaker, A., & El-Rabbany, A., 2017. Multispectral LiDAR Data for Land Cover Classification of Urban Areas. Sensors, 17, 958. https://doi.org/10.3390/s17050958.

Nachtergaele, F. O. F., & Licona-Manzur, C., 2008. The Land Degradation Assessment in Drylands (LADA) Project: Reflections on Indicators for Land Degradation Assessment.

National Land Survey of Finland, (n.d.). Laser scanning data 5 p. National Land Survey of Finland. https://www.maanmittauslaitos.fi/en/maps-and-spatial-data/datasets-and-interfaces/product-descriptions/laser-scanning-data-5-p (accessed Oct 26, 2025).

Pan, S., Guan, H., Chen, Y., Yu, Y., Nunes Gonçalves, W., Marcato Junior, J., & Li, J., 2020. Land-cover classification of multispectral LiDAR data using CNN with optimized hyper-parameters. ISPRS J. Photogramm. Remote Sens., 166, 241–254. https://doi.org/10.1016/j.isprsjprs.2020.05.022.

Pan, S., Guan, H., Yu, Y., Li, J., & Peng, D., 2019. A comparative land-cover classification feature study of learning algorithms: DBM, PCA, and RF using multispectral LiDAR data. IEEE J. Sel. Top. Appl. Earth Obs. Remote. Sens., 12, 1314–1326.

Pfeifer, N., Mandlburger, G., Otepka, J., & Karel, W., 2014. OPALS – a framework for airborne laser scanning data analysis. Comput. Environ. Urban. Syst., 45, 125–136.

Pfennigbauer, M., & Ullrich, A., 2010. Improving quality of laser scanning data acquisition through calibrated amplitude and pulse deviation measurement. Proc. SPIE, 7684. https://doi.org/10.1117/12.849641.

Qi, C.R., Su, H., Mo, K., & Guibas, L.J., 2017a. PointNet: Deep learning on point sets for 3D classification and segmentation. Proc. IEEE Conf. Comput. Vis. Pattern Recognit. (CVPR), 652–660. https://doi.org/10.1109/CVPR.2017.16.

Qi, C.R., Yi, L., Su, H., Guibas, L.J., 2017b: PointNet++: Deep hierarchical feature learning on point sets in a metric space. Proc. 31st Int. Conf. Neural Inf. Process. Syst. (NIPS'17), Curran Associates Inc., Red Hook, NY, USA, 5105–5114.

QingWwang, WQ., & Zhang, Z., Gu, Y., 2024. Multi-Kernel Graph Structure Learning for Multispectral Point Cloud Classification. IEEE J. Sel. Top. Appl. Earth Obs. Remote Sens., PP, 1–15. https://doi.org/10.1109/JSTARS.2024.3368472.

Riveiro, B., Lindenbergh, R., 2019. *Laser Scanning: An Emerging Technology in Structural Engineering*. CRC Press, Boca Raton, FL, USA, Vol. 14.

Rizaldy, A., Gloaguen, R., Fassnacht, F. E., & Ghamisi, P., 2025. HyperPointFormer: Multimodal Fusion in 3-D Space With Dual-Branch Cross-Attention Transformers. IEEE J. Sel. Top. Appl. Earth Obs. Remote Sens., 18, 21254–21274.

Robert, D., Raguet, H., & Landrieu, L., 2023. Efficient 3D semantic segmentation with Superpoint Transformer. Proc. IEEE/CVF Int. Conf. Comput. Vis. (ICCV), 1785–1795.

Ruoppa, L., Hietala, T., Seppänen, V., Taher, J., Hakala, T., Yu, X., Kukko, A., Kaartinen, H., & Hyyppä, J., 2026. Benchmarking individual tree segmentation using multispectral airborne laser scanning data: The FGI-EMIT dataset. ISPRS Journal of Photogrammetry and Remote Sensing, 236, 569–605. https://doi.org/10.1016/j.isprsjprs.2026.04.021.

Ruoppa, L., Oinonen, O., Taher, J., Lehtomäki, M., Takhtkeshha, N., Kukko, A., Kaartinen, H., Hyyppä, J., 2025. Unsupervised deep learning for semantic segmentation of multispectral LiDAR forest point clouds. ISPRS J. Photogramm. Remote Sens. 228, 694–722. https://doi.org/10.1016/j.isprsjprs.2025.07.038.

Saurabh Prasad, Bertrand Le Saux, Naoto Yokoya, & Ronny Hansch, 2020. 2018 IEEE GRSS Data Fusion Challenge – Fusion of Multispectral LiDAR and Hyperspectral Data. IEEE Dataport. https://doi.org/10.21227/jnh9-nz89.

Serna, A., Marcotegui, B., Goulette, F., & Deschaud, J.-E., 2014. Paris-rue-Madame database: a 3D mobile laser scanner dataset for benchmarking urban detection, segmentation and classification methods. ICPRAM 2014, 1–4.

Taher, J., Hakala, T., Jaakkola, A., Hyyti, H., Kukko, A., Manninen, P., Maanpää, J., Hyyppä, J., 2022. Feasibility of Hyperspectral Single Photon Lidar for Robust Autonomous Vehicle Perception. Sensors 22, 5759. https://doi.org/10.3390/s22155759.

Taher, J., Hyyppä, E., Hyyppä, M., Salolahti, K., Yu, X., Matikainen, L., Kukko, A., Lehtomäki, M., Kaartinen, H., Thurachen, S., et al., 2026. Multispectral airborne laser scanning for tree species classification: A benchmark of machine learning and deep learning algorithms. ISPRS J. Photogramm. Remote Sens. 233, 278–309. https://doi.org/10.1016/j.isprsjprs.2026.01.031.

Takhtkeshha, N., Bayrak, O.C., Mandlburger, G., Remondino, F., Kukko, A., Hyyppä, J.,2024a. Automatic annotation of 3D multispectral LiDAR data for land cover classification. In: IGARSS 2024 - 2024 IEEE International Geoscience and Remote Sensing Symposium. pp. 8645–8649. http://dx.doi.org/10.1109/IGARSS53475.2024.10642907.

Takhtkeshha, N., Bocaux, L., Ruoppa, L., Remondino, F., Mandlburger, G., Kukko, A., & Hyyppä, J., 2025a. 3D forest semantic segmentation using multispectral LiDAR and 3D deep learning. arXiv. https://doi.org/10.48550/arXiv.2507.08025.

Takhtkeshha, N., Mandlburger, G., Remondino, F., & Hyyppä, J., 2024b. Multispectral light detection and ranging technology and applications: A review. Sensors, 24(5), 1669. https://doi.org/10.3390/s24051669.

Takhtkeshha, N., Mazzacca, G., Remondino, F., Hyyppä, J., & Mandlburger, G., 2025b. Multispectral LiDAR data for extracting tree points in urban and suburban areas. https://doi.org/10.48550/arXiv.2508.19881.

Thomas, H., Qi, C. R., Deschaud, J.-E., Marcotegui, B., Goulette, F., Guibas, L., 2019. KPConv: Flexible and deformable convolution for point clouds. In: *2019 IEEE/CVF International Conference on Computer Vision (ICCV)*, Seoul, South Korea, pp. 6410–6419. https://doi.org/10.1109/ICCV.2019.00651.

Thomas, H., Tsai, Y.-H.H., Barfoot, T.D., Zhang, J., 2024. KPConvX: Modernizing Kernel Point Convolution with Kernel Attention. Proc. IEEE/CVF Conf. Comput. Vis. Pattern Recognit. 5525–5535.

Tychola, K. A., Vrochidou, E., & Papakostas, G. A., 2024. Deep learning based computer vision under the prism of 3D point clouds: a systematic review. *The Visual Computer*, 40(11), 8287–8329. https://doi.org/10.1007/s00371-023-03237-7.

Varney, N., Asari, V. K., & Graehling, Q., 2020. DALES: A large-scale aerial LiDAR data set for semantic segmentation. In *Proceedings of the IEEE/CVF Conference on Computer Vision and Pattern Recognition Workshops*.

Walicka, A., Pfeifer, N., 2025. Standard classes for urban topographic mapping with ALS: classification scheme and a first implementation. Remote Sens. 17, 2731. https://doi.org/10.3390/rs17152731.

Wang, Q., Chen, X., Zhang, Z., Meng, Y., Gu, Y., 2025a. Masking graph cross-convolution network for multispectral point cloud classification. IEEE Trans. Geosci. Remote Sens. PP, 1–1. https://doi.org/10.1109/TGRS.2025.3545783.

Wang, X., Song, L., Feng, Y., & Zhu, J., 2025b. S3F2Net: Spatial-Spectral-Structural Feature Fusion Network for Hyperspectral Image and LiDAR Data Classification. IEEE Transactions on Circuits and Systems for Video Technology, PP. 1–1. https://doi.org/10.1109/TCSVT.2025.3525734.

Wang, Y., Sun, Y., Cao, X., Wang, Y., Zhang, W., & Cheng, X., 2023. A review of regional and global scale Land Use/Land Cover (LULC) mapping products generated from satellite remote sensing. ISPRS J. Photogramm. Remote Sens., 206, 311–334. https://doi.org/10.1016/j.isprsjprs.2023.11.014.

Wang, Y., Sun, Y., Liu, Z., Sarma, S. E., Bronstein, M. M., Solomon, J. M., 2019. Dynamic Graph CNN for learning on point clouds. *ACM Trans. Graph. (TOG)*.

Wu, L., Chen, Y., Le, Y., Qian, Y., Zhang, D., Wang, L., 2024a. A high-precision fusion bathymetry of multi-channel waveform curvature for bathymetric LiDAR systems. Int. J. Appl. Earth Obs. Geoinf. 128, 103770. https://doi.org/10.1016/j.jag.2024.103770.

Wu, X., Jiang, L., Wang, P.-S., Liu, Z., Liu, X., Qiao, Y., Ouyang, W., He, T., Zhao, H., 2024b. Point Transformer V3: Simpler, Faster, Stronger. Proc. IEEE/CVF Conf. Comput. Vis. Pattern Recognit. (CVPR).

Xiao, K., Qian, J., & Li, T., 2022. Multispectral LiDAR point cloud segmentation for land cover leveraging semantic fusion in deep learning network. Remote Sens., 15, 243.

Yang, J., Gan, R., Luo, B., Wang, A., Shi, S., & Du, L., 2024. An improved method for individual tree segmentation in complex urban scene based on using multispectral LiDAR by deep learning. IEEE J. Sel. Top. Appl. Earth Obs. Remote Sens., 1–17.

Zhang, Z., Li, T., Tang, X., Lei, X., & Peng, Y., 2022. Introducing improved transformer to land cover classification using multispectral LiDAR point clouds. Remote Sens., 14, 3808.

Zhang, Z., Yang, B., Wang, B., & Li, B., 2023. GrowSP: Unsupervised semantic segmentation of 3D point clouds. In 2023 IEEE/CVF Conference on Computer Vision and Pattern Recognition, 17619–17629. https://doi.org/10.1109/CVPR52729.2023.01690.

Zhao, H., Jiang, L., Jia, J., Torr, P. H. S., Koltun, V., 2021a. Point Transformer. In: *2021 IEEE/CVF Conference on Computer Vision and Pattern Recognition (CVPR)*, pp. 16259–16268. https://doi.org/10.1109/CVPR46437.2021.01595.

Zhao, P., Guan, H., Li, D., Yu, Y., Wang, H., Gao, K., Marcato Junior, J., & Li, J., 2021b. Airborne multispectral LiDAR point cloud classification with a feature reasoning-based graph convolution network. Int. J. Appl. Earth Obs. Geoinf., 105, 102634.

Zhou, G., Qi, H., Shi, S., Bi, S., Tang, X., Gong, W., 2025a. Spatial–spectral feature fusion and spectral reconstruction of multispectral LiDAR point clouds by attention mechanism. Remote Sens. 17, 2411. https://doi.org/10.3390/rs17142411.

Zhou, W., Liu, K., Jin, W., Wang, Q., She, Y., Yu, Y., Ma, C., 2025b. Advancements in deep learning for point cloud classification and segmentation: A comprehensive review. Comput. Graph. 130, 104238. https://doi.org/10.1016/j.cag.2025.104238.

Zolanvari, I., Ruano, S., Rana, A., Cummins, A., Smolic, A., Da Silva, R., & Rahbar, M., 2019. DublinCity: Annotated LiDAR Point Cloud and its Applications.